\documentclass[letterpaper]{article} %
\usepackage{aaai2026}  %
\usepackage{times}  %
\usepackage{helvet}  %
\usepackage{courier}  %
\usepackage[hyphens]{url}  %
\usepackage{graphicx} %

\usepackage{booktabs}
\usepackage{multirow}
\usepackage{subcaption}
\usepackage{amsmath}
\usepackage{amssymb}
\usepackage{cleveref}
\usepackage{tikz}
\usepackage{amsthm}
\usepackage{comment}

\theoremstyle{plain}
\newtheorem{theorem}{Theorem}

\theoremstyle{definition}
\newtheorem{definition}[theorem]{Definition}

\theoremstyle{remark}

\usepackage{natbib}  %
\usepackage{caption} %
\usepackage{algorithm}
\usepackage{algorithmic}

\usepackage{newfloat}
\usepackage{listings}
\DeclareCaptionStyle{ruled}{labelfont=normalfont,labelsep=colon,strut=off} %
\floatstyle{ruled}
\newfloat{listing}{tb}{lst}{}
\floatname{listing}{Listing}
\title{PeriUn: Enhancing Unlearning by Selectively Forgetting Peripheral Samples}
\author {
    HEE BIN YOO\textsuperscript{\rm 1},
    Dong-Sig Han\textsuperscript{\rm 2},
    Jaein Kim\textsuperscript{\rm 1},
    Byoung-Tak Zhang\textsuperscript{\rm 1}\thanks{Corresponding author}
}
\affiliations {
    \textsuperscript{\rm 1}Seoul National University, \\
    \textsuperscript{\rm 2}Imperial College London\\
    yooheebin@snu.ac.kr, ehd4585@gmail.com, qpwodlsqp@snu.ac.kr,  btzhang@snu.ac.kr
}

\usepackage{bibentry}

\begin{document}

\maketitle

\begin{abstract}
Once trained, neural networks memorize information in diffusely encoded parameters, making it difficult to forget in support of the right to be forgotten.
Unlearning aims to remove the influence of data, with performance measured against a retrained model that excludes the data.
However, understanding the behavior of gold-standard retraining remains underexplored.
We compare original and retrained models and observe that most prediction changes occur in peripheral samples near decision boundaries.
Consequently, we propose PeriUn, a selective strategy that unlearns only peripheral samples to mimic retrained model behavior with minimal disruption, unlike prior works that remove the entire request.
Combined with the Random Label based method, PeriUn significantly improves both generalization and privacy metrics. Specifically, on TinyImageNet with VGG16, PeriUn increases the Tug-of-War score by 22 points compared to the strongest. 
Besides, the MIA gap score surpasses the state-of-the-art method, improving by 8.7 points after applying PeriUn.
Further analyses confirm that PeriUn better preserves the feature space and aligns closely with the retrained model.
\end{abstract}

\section{Introduction}
With growing concerns over privacy regulations and the ``right to be forgotten'' \citep{mantelero2013eu, cl2017ab}, unlearning techniques designed to remove the influence of specific data from networks have garnered increasing attention \citep{cao15towards}.
As an ideal solution, a retrained-from-scratch model trained only on the retain set, i.e., samples not subject to deletion requests, serves as the oracle, but it is prohibitively expensive to obtain.
Therefore, recent studies \citep{triantafillou2024makingprogressunlearningfindings, ebrahimpourboroojeny2025amunadversarialmachineunlearning} have aimed to make unlearning methods approximate the accuracy of retraining.
For instance, various studies have proposed to encourage sparsity \citep{jia2024modelsparsitysimplifymachine} or preserve non-salient parameters \citep{fan2024salunempoweringmachineunlearning}.
However, approximation based approaches have inherent limitations, as they may unintentionally remove useful information from the retain set, i.e., catastrophic forgetting, an issue addressed by \citet{he2024naturalmachineunlearning, chen2024machineunlearningnullspace, peng2025adversarialmixupunlearning} through explicit retain set information preservation.

\begin{figure}[t]
    \centering
    \begin{subfigure}{0.98\columnwidth}
        \centering
        \begin{tikzpicture}
            \node[inner sep=0] (img) {\includegraphics[width=0.95\linewidth]{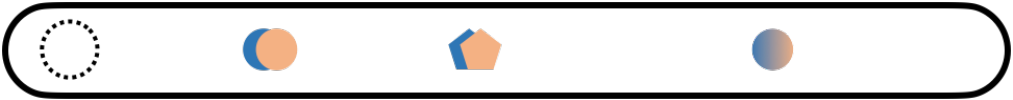}};
           \node[font=\scriptsize, text=black] at (-2.61,0) {Periphery};
           \node[font=\scriptsize, text=black] at (-1.05,0) {Forget Set};
           \node[font=\scriptsize, text=black] at (0.905,0.01) {Class 1\&2 Center};
           \node[font=\scriptsize, text=black] at (2.94,0) {Pred. Change};
        \end{tikzpicture}
    \end{subfigure}
    \begin{subfigure}{0.85\columnwidth}
        \centering
        \begin{tikzpicture}
            \node[inner sep=0] (img) {\includegraphics[width=0.95\linewidth]{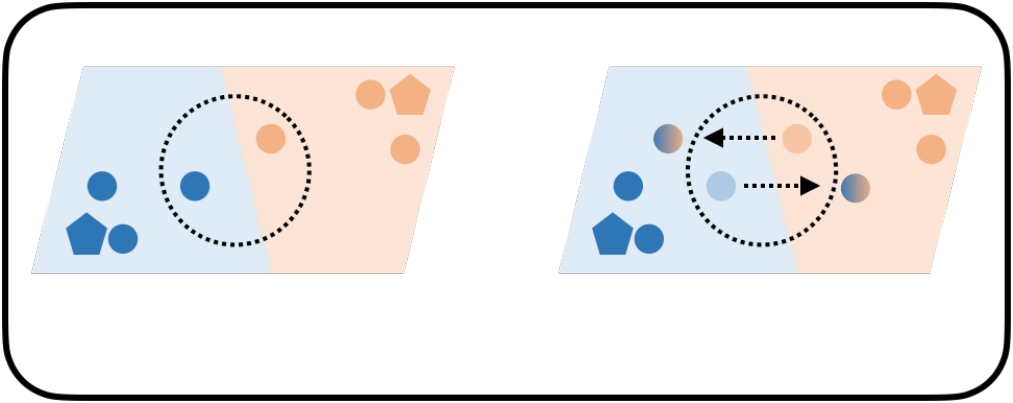}};
           \node[font=\large, text=black] at (-1.78,-0.85) {$\mathcal{M}_{o}$};
           \node[font=\large, text=black] at (1.65,-0.85) {$\mathcal{M}_{r}$};
        \end{tikzpicture}
        \caption{Discrepancy analysis}
    \end{subfigure}
    \begin{subfigure}{0.85\columnwidth}
        \centering
        \begin{tikzpicture}
            \node[inner sep=0] (img) {\includegraphics[width=0.95\linewidth]{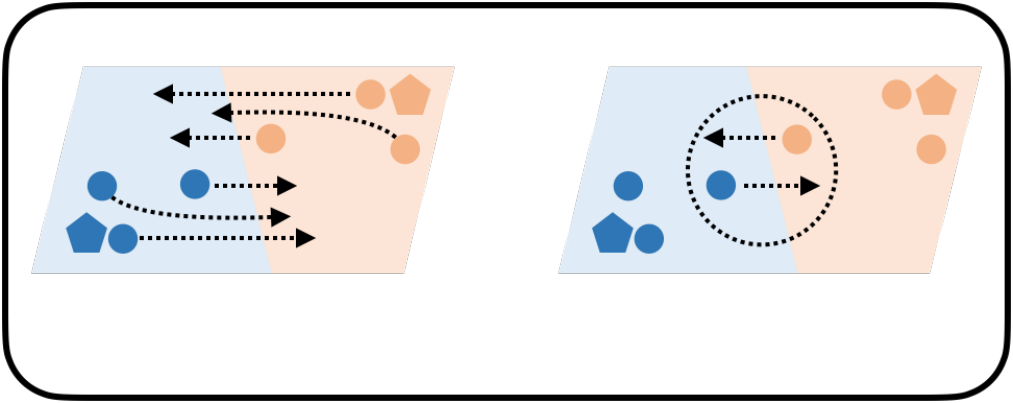}};
            \node[font=\large, text=black] at (-1.78,-0.85) {Baseline};
            \node[font=\large, text=black] at (1.65,-0.85) {Baseline $+$ PeriUn};
        \end{tikzpicture}
        \caption{Proposed method PeriUn}
        \label{fig:main_method}
    \end{subfigure}
    \caption{(Top) Discrepancy observed between original model $\mathcal{M}_o$ and retrained model $\mathcal{M}_r$, where peripheral samples predominantly contribute to confidence and prediction changes. (Bottom) Comparison between PeriUn and the standard baseline.
    PeriUn is a selective unlearning strategy that provides only the periphery samples in the forget set to the unlearning algorithm.
    This approach mimics the behavior of retraining, thereby mitigating catastrophic forgetting.
    }
    \label{fig:main}
\end{figure}

Notably, catastrophic forgetting does not occur in the oracle retrained model.
This observation motivates our central arguments; understanding and mimicking the behavior of the retrained model can guide more effective unlearning methods that mitigate catastrophic forgetting and improve performance. 
To explore these arguments, we investigate the following research questions:
\begin{quote}
\emph{What are the differences between the original and retrained models, and their causes? Can these differences be leveraged to enhance unlearning?}
\end{quote}

\noindent To investigate these questions, we analyze changes in terms of confidence and prediction correctness of forget set samples by comparing the original and retrained models.
In the original model, these samples tend to be predicted with overconfidence and near-perfect accuracy, whereas both of their confidence and accuracy noticeably decline in the retrained model.
To identify the causes responsible of this shift, we conduct additional analyses at the individual sample level.
A notable finding is that predictions change from correct to incorrect with reduced confidence, and this occurs primarily in peripheral samples near decision boundaries.
This directly addresses our first research question.
In addition, our fine-grained analysis extends prior work \citep{seo2025revisit, georgiev2025machine}, which focuses solely on macroscopic changes between the original and retrained models.

Based on this observation, we hypothesize that selectively unlearning on peripheral samples rather than the entire set can effectively prevent catastrophic forgetting and improve performance.
The above hypothesis guides our approach to the second research question.
We propose Peripheral Unlearning, \textbf{PeriUn}, a selective unlearning strategy that targets only the peripheral samples within the forget set (see \Cref{fig:main}), mimicking retrained model by focusing on samples that mainly cause changes in confidence and prediction.

Using PeriUn, we conducted comparative experiments with various leading unlearning methods across diverse architectures and datasets, evaluating their ability to mimic retraining in terms of performance, privacy, and prediction behavior.
In particular, combining PeriUn with the Saliency Unlearn \citep{fan2024salunempoweringmachineunlearning} led to significant improvements.
For instance, on the TinyImageNet dataset with the VGG16-bn model, the ToW score improved by 22.04 points on a 100-point scale compared to the best baseline. 
Moreover, the MIA gap score surpasses the previous best method, improving by 8.7 points relative to baselines without PeriUn.
In addition, we quantitatively and qualitatively analyzed the prediction showing PeriUn enables the unlearned model to better mimic the retrained model compared to the baseline, while reducing catastrophic forgetting. Furthermore, we validated the effectiveness of PeriUn through ablations.

The main contributions of this work are as follows:
\begin{itemize}
    \item Our work presents a novel analysis of how individual forget-set samples change from the original to retrained models, uncovering patterns for effective unlearning.
    \item Building on these insights, we introduce a selective forgetting strategy that closely approximates the retrained model and mitigates catastrophic forgetting.
    \item Extensive experiments on image classification benchmarks demonstrate state-of-the-art ToW and MIA gap performance of PeriUn, validating its effectiveness.
\end{itemize}

\section{Related Works}
\paragraph{Coreset.}
Existing studies demonstrating the existence of coresets in neural network training inspire our selective forgetting approach.
Notably, \citet{toneva2019empiricalstudyexampleforgetting, paul2023deeplearningdatadiet} demonstrated that comparable results can be achieved without utilizing the entire training dataset. %
Similarly, \citet{guo2022deepcorecomprehensivelibrarycoreset} constructed coresets based on confidence-based selection, aligning with our approach.%
In addition, \citet{jain2022databasedperspectivetransferlearning, zhang2023selectivitydrivesproductivityefficient} confirmed the presence of coresets in transfer learning scenarios.
Collectively, these studies on coresets suggest that, in the context of unlearning, it may be sufficient to selectively remove a few samples to mimic the retrained model, rather than indiscriminately deleting all data.

\paragraph{Memorization.}
Prior research on memorization supports both the validity and necessity of the Peripheral Unlearning strategy proposed in this work.
\citet{carlini2019prototypical, jiang2021char} empirically demonstrated that while some samples are correctly predicted even when excluded from training, others make incorrect predictions when held out. Notably, \citet{jiang2021char} observed that such vulnerable samples tend to be visually ambiguous and have lower average confidence. %
These findings align with \citet{feldman2020neuralnetworksmemorizewhy, feldman2021doeslearningrequirememorization}, which showed that samples susceptible to incorrect prediction under hold-out are \textit{memorized} samples and memorization contributes to model performance.
Taken together, the above results suggest that it is more principled to selectively remove \textit{memorized} samples supporting the strategy of PeriUn.

\paragraph{Unlearning.}
There have been diverse approaches to erase the influence of data to be deleted on the trained model.
\textit{Fine-Tuning (FT)} tunes the model using only the retain set, thereby inducing catastrophic forgetting of the forget set \citep{golatkar2020eternalsunshinespotlessnet}. %
\textit{Gradient Ascent (GA)} updates the model in the direction that maximizes the loss on the forget set, encouraging removal of the corresponding information \citep{thudi2022unrollingsgdunderstandingfactors}. %
\textit{Random Label (RL)} perturbs the learned representation by assigning random labels to the forget set and fine-tuning the model on the entire dataset \citep{hayase2020selectiveforgettingdeepnetworks}. %
\textit{Gradient Ascent–Gradient Descent (GAGD)} applies GA to the forget set and GD to the retain set, aiming to balance removal and preservation \citep{kurmanji2023unboundedmachineunlearning}.
\textit{Fine-Tuning with L1 Regularization (FT$+$L1)} encourages parameter sparsity through L1 regularization, improving unlearning effectiveness via pruning effects \citep{jia2024modelsparsitysimplifymachine}.
\textit{Saliency Unlearning (SU)} performs RL-style unlearning by selectively updating only salient parameters based on gradient analysis \citep{fan2024salunempoweringmachineunlearning}.
Additional methods are in the appendix.

\section{Problem Setup}
We consider the problem of machine unlearning in image classification \citep{zhao2024makesunlearninghard, fan2024salunempoweringmachineunlearning}. In particular, we focus on handling random instance removal requests to comply with user data deletion demands \citep{golatkar2021mixedprivacyforgettingdeepnetworks, ijcai2021p137, he2021deepobliviatepowerfulcharmerasing}.
Given a training dataset $\mathcal{D}_{\text{train}}$, test dataset $\mathcal{D}_{t}$, and an original model $\mathcal{M}_{o}$ trained on $\mathcal{D}_{\text{train}}$, a user may request to remove the influence of a specific subset $\mathcal{D}_f \subset \mathcal{D}_{\text{train}}$, referred to as the forget set.

\noindent \textbf{Approximate Unlearning.}\enspace The ideal way to fulfill this request is to retrain a model from scratch on the remaining data $\mathcal{D}_r = \mathcal{D}_{\text{train}} \setminus \mathcal{D}_f$, resulting in a retrained model $\mathcal{M}_{r}$. This process, referred to as exact unlearning, produces an oracle model for comparison.
However, retraining from scratch incurs computational costs equivalent to training $\mathcal{M}_{o}$, which is often prohibitively expensive in practice.
As a result, approximate unlearning methods aim to transform $\mathcal{M}_{o}$ into an unlearned model $\mathcal{M}_{u}$ without full retraining. These methods are given access to $\mathcal{M}_{o}$, the forget set $\mathcal{D}_f$, and the retain set $\mathcal{D}_r$, and seek to approximate the behavior of $\mathcal{M}_{r}$.

\noindent \textbf{Evaluation Metric.}\enspace To evaluate the quality of unlearning, we adopt the Tug-of-War (ToW) and MIA gap metric introduced by \citet{zhao2024makesunlearninghard}.
ToW evaluates how well $\mathcal{M}_{u}$ mimics the predictive behavior of $\mathcal{M}_{r}$. At the same time, the MIA gap captures the similarity in membership inference attack success rates, reflecting alignment in privacy behavior.

\begin{definition}[Tug-of-War]
Let $a^m_s$ denote the classification accuracy of model $\mathcal{M}_m$ on dataset $\mathcal{D}_s$. The ToW score between $\mathcal{M}_{u}$ and $\mathcal{M}_{r}$ is defined as:
\begin{equation}
\begin{aligned}
    &\text{ToW}(\mathcal{M}_r, \mathcal{M}_u, \mathcal{D}_f, \mathcal{D}_r, \mathcal{D}_t) \\
    & = (1 - |a^{r}_f - a^{u}_f|) \cdot 
           (1 - |a^{r}_r - a^{u}_r|) \cdot 
           (1 - |a^{r}_t - a^{u}_t|) \cdot 100.
\end{aligned}
\end{equation}
\end{definition}
\noindent A ToW score closer to 100 indicates a better approximation of the retrained model’s behavior across all data splits.

\noindent \textbf{Experimental Setup.}\enspace We evaluate unlearning performance on four dataset–architecture pairs. Specifically, we use CIFAR10, CIFAR100~\citep{krizhevsky09learningmultiple}, and TinyImageNet~\cite{le2015tiny} with ResNet18, ResNet50~\citep{he2015deepresiduallearningimage}, and VGG16-bn~\cite{simonyan2015deepconvolutionalnetworkslargescale} following \citet{zhao2024makesunlearninghard}. 
Each row in \Cref{tab:orig_model_perf} corresponds to \textit{Setup 1–4} and reports the accuracy of $\mathcal{M}_o$.
For statistical significance, we use three different fixed random initializations. 
Then, we generate and fix three random forget sets of size 3000.
We utilize an NVIDIA RTX 3090, an Intel Xeon 5218, and 128 GB of memory with Ubuntu 18.04.

\begin{table}[t]
        \centering
        {\fontsize{9}{11}\selectfont
        \begin{tabular}{cc|cc}
            \hline
             & Dataset - Model    & Train          & Test           \\ \hline
            1 & CIFAR10 - ResNet18 & 99.46$\pm$0.11 & 92.31$\pm$0.42 \\
            2 & CIFAR100 - ResNet50 & 99.97$\pm$0.01 & 76.32$\pm$0.26 \\
            3 & TinyImagenet - ResNet18 & 99.98$\pm$0.00 & 65.19$\pm$0.35 \\
            4 & TinyImagenet - VGG16-bn & 99.98$\pm$0.00 & 60.40$\pm$0.59 \\ \hline
        \end{tabular}
        }
        \captionof{table}{Train/test classification accuracy of the original models $\mathcal{M}_o$ used in the unlearning experiments. The networks classify the training data with near-perfect accuracy.}
        \label{tab:orig_model_perf}
\end{table}

\section{Preliminary Analysis of the Retrained Model}
To address the first question, we analyze output differences between \(\mathcal{M}_o\) and \(\mathcal{M}_r\), especially on \(\mathcal{D}_f\), and examine how these results can be leveraged to improve unlearning.

\noindent \textbf{Observation 1.}\enspace 
We demonstrate that the output prediction of data subsets ($\mathcal{D}_r, \mathcal{D}_f, \mathcal{D}_t$) drawn from the same underlying distribution, split by i.i.d.\ sampling, can be separated based on whether the data were used during training or not. %

\noindent \textbf{Observation 2.}  \enspace
Building on the above observation and considering that \(\mathcal{D}_f\) is used for training in \(\mathcal{M}_o\) but not in \(\mathcal{M}_r\), we analyze what individual samples in \(\mathcal{D}_f\) change to exhibit distinct characteristics as seen and unseen in train.

\begin{figure}[t]
	\centering
	\begin{subfigure}[b]{0.485\columnwidth}
		\centering
		\includegraphics[width=\linewidth]{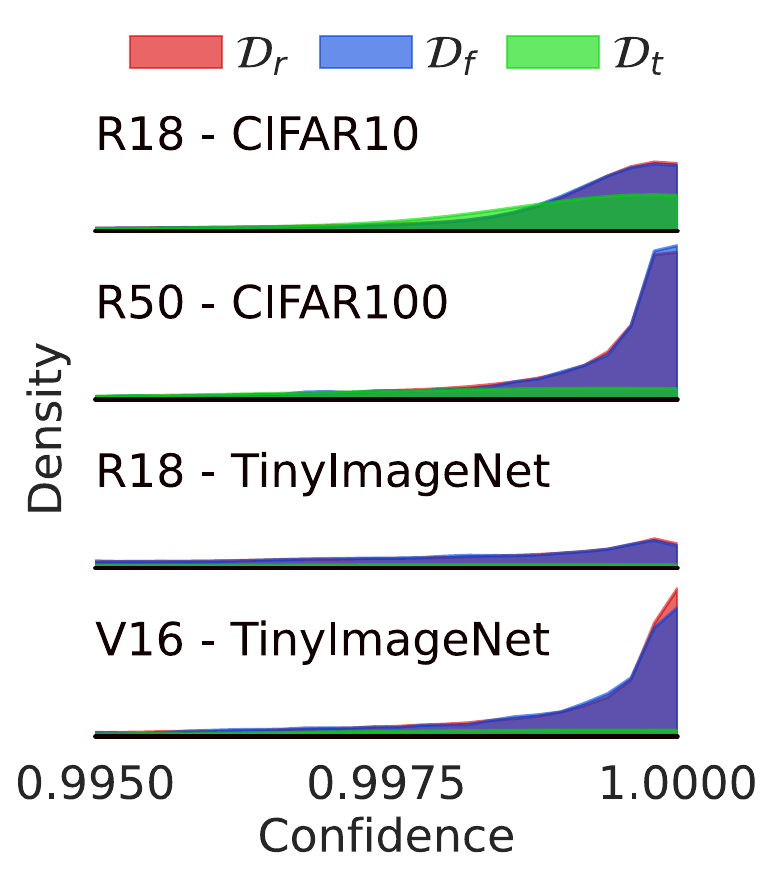}
		\caption{$\mathcal{M}_o$}
	\end{subfigure}
	\begin{subfigure}[b]{0.485\columnwidth}
		\centering
		\includegraphics[width=\linewidth]{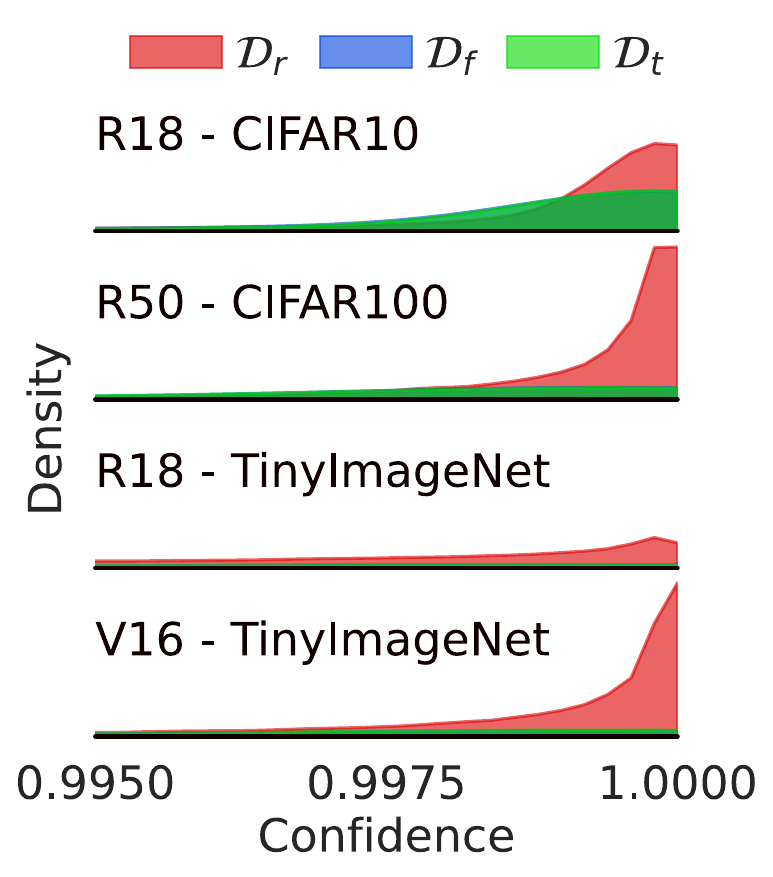}
		\caption{$\mathcal{M}_r$}
	\end{subfigure}
	\caption{
    Observation 1: Seen vs. Unseen confidence analysis. 
    Seen data refers to $(\mathcal{D}_r, \mathcal{D}_f)$ under $\mathcal{M}_o$, as well as $\mathcal{D}_r$ under $\mathcal{M}_r$, which are used in training, while all others are considered unseen. We observe that the model is overconfident in seen data, while exhibiting low confidence in unseen. Note that in (b), $\mathcal{D}_f$ and $\mathcal{D}_t$ overlap.}
	\label{fig:confidence-analysis}
\end{figure}

\subsection{Observation 1: Seen vs. Unseen}
We observe that the confidence distributions are distinguishable concerning whether \(\mathcal{D}_r\), \(\mathcal{D}_f\), or \(\mathcal{D}_t\) was used during training, and samples within each data subset share common internal characteristics.  
Note that we define confidence as the maximum softmax probability across all classes.
We measure the confidence density of \(\mathcal{M}_o\) and \(\mathcal{M}_r\) trained under four different setups and visualize the results of the distributions. In detail, we subsample each set to match the size of \(\lvert \mathcal{D}_f \rvert\) for a fair comparison and plot the histogram’s KDE.

Our experimental results in \Cref{fig:confidence-analysis} show that samples in the seen subsets tend to exhibit overconfidence, while those in the unseen set generally show underconfidence.
This observation is consistent with findings from prior studies \citep{ shokri2017membershipinferenceattacksmachine, %
datta2025topologyoutofdistributionexamplesdeep}.  
We also found consistent differences in accuracy between seen and unseen data, regardless of data subsets or random seed.
As shown in \Cref{tab:orig_model_perf} and \Cref{tab:comparision}, the average accuracy on seen datasets is always near 100\% within one percentage point. In comparison, accuracy differences among unseen datasets remain within one percentage point.  
These results confirm that the datasets \(\mathcal{D}_{r}\), \(\mathcal{D}_{f}\), and \(\mathcal{D}_t\), which are separated via i.i.d. sampling, can be divided into two groups based on their training exposure, i.e., seen vs. unseen, and that each group shares nearly identical accuracy and confidence distribution patterns. Additional results are in the appendix \emph{Confidence Distribution Results}.

\subsection{Observation 2: Discrepancy Analysis on $\mathcal{D}_f$}
The forget dataset \(\mathcal{D}_f\) is seen by \(\mathcal{M}_o\) but unseen by \(\mathcal{M}_r\), leading \(\mathcal{M}_o\) to show overconfident, near-perfect accuracy, while \(\mathcal{M}_r\) has lower confidence and accuracy on \(\mathcal{D}_f\).
To understand what drives this difference at the level of individual samples, we perform an analysis to identify samples that contribute to the shift from seen to unseen behavior.

\begin{figure}[t]
	\centering
	\begin{subfigure}[b]{0.9\columnwidth}
		\centering
		\includegraphics[width=\linewidth]{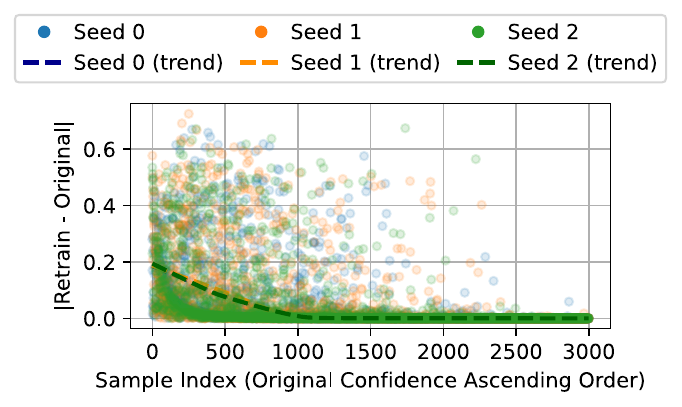}
		\caption{Confidence difference between $\mathcal{M}_o$ and $\mathcal{M}_r$ on $\mathcal{D}_f$}
		\label{fig:conf-diff}
	\end{subfigure}
	\begin{subfigure}[b]{0.9\columnwidth}
		\centering
		\includegraphics[width=\linewidth]{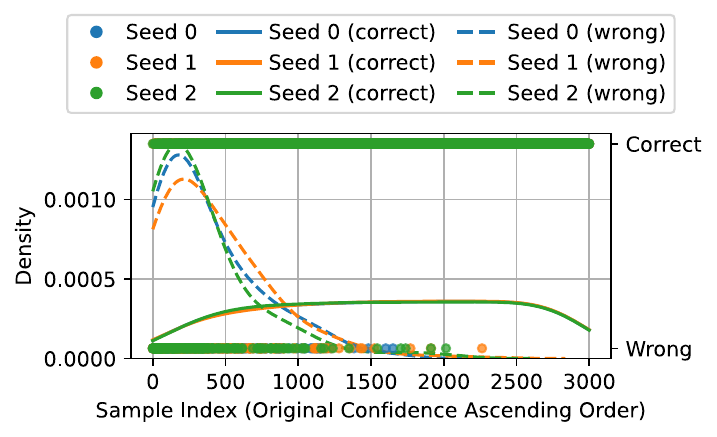}
		\caption{Correctness density of $\mathcal{M}_r$ on $\mathcal{D}_f$.}
		\label{fig:correctness}
	\end{subfigure}
	\caption{Discrepancy analysis on $\mathcal{D}_f$ in Setup 1. %
    Low-confidence samples exhibit greater confidence shifts and are the most misclassified samples in $\mathcal{M}_r$.}
	\label{fig:conf-diffandcorr}
\end{figure}

\begin{figure}[t]
	\centering
    \begin{subfigure}[b]{\columnwidth}
		\centering
		\includegraphics[width=0.5\linewidth]{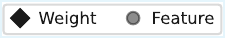}
	\end{subfigure}
	\begin{subfigure}[b]{0.485\columnwidth}
		\centering
		\includegraphics[width=\linewidth]{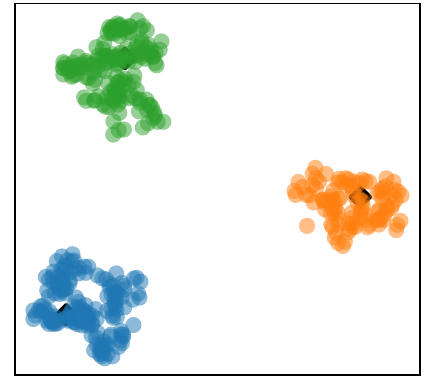}
		\caption{High confidence samples in $\mathcal{M}_o$ feature space.}
        \label{fig:main_tsne_a}
	\end{subfigure}
	\begin{subfigure}[b]{0.485\columnwidth}
		\centering
		\includegraphics[width=\linewidth]{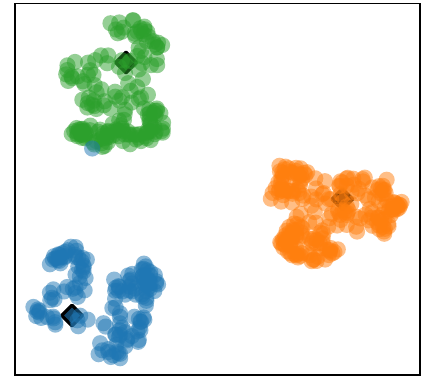}
		\caption{Low confidence samples in $\mathcal{M}_o$ feature space.}
        \label{fig:main_tsne_b}
	\end{subfigure}
	\begin{subfigure}[b]{0.485\columnwidth}
		\centering
		\includegraphics[width=\linewidth]{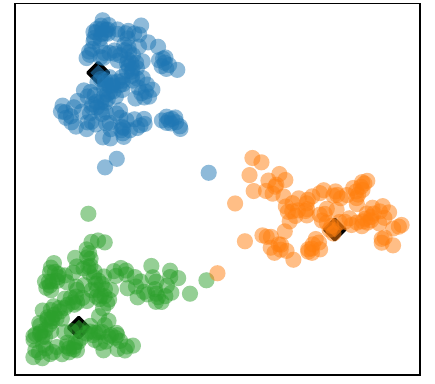}
		\caption{High confidence samples in $\mathcal{M}_r$ feature space.}
        \label{fig:main_tsne_c}
	\end{subfigure}
	\begin{subfigure}[b]{0.485\columnwidth}
		\centering
		\includegraphics[width=\linewidth]{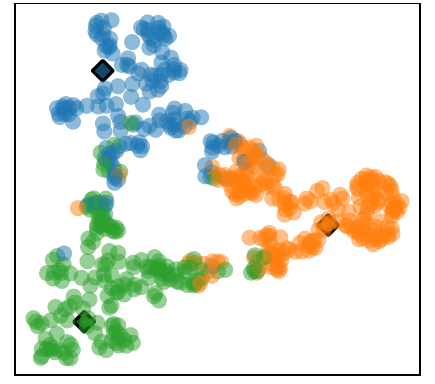}
		\caption{Low confidence samples $\mathcal{M}_r$ feature space.}
        \label{fig:main_tsne_d}
	\end{subfigure}
	\caption{T-SNE of $\mathcal{D}_f$ features from $\mathcal{M}_o$ and $\mathcal{M}_r$ in Setup 1. Samples are split by the median $\mathcal{M}_o$ confidence. %
    }
	\label{fig:tsne}
\end{figure}

\noindent \textbf{Confidence Difference Analysis.}  \enspace
We measured the confidence of each sample in the forget set \(\mathcal{D}_f\) using both \(\mathcal{M}_o\) and \(\mathcal{M}_r\), then compared their differences.  
As in \Cref{fig:conf-diff}, samples were sorted in ascending order based on their confidence from \(\mathcal{M}_o\), and Locally Weighted Scatterplot Smoothing was applied to visualize the overall trend.  
High-confidence samples in \(\mathcal{M}_o\) showed little change and, thus, remained confident in \(\mathcal{M}_r\), while lower-confidence samples exhibited larger changes.
This suggests that less confident samples are more susceptible to retraining.
This result can be interpreted from the perspective of the Neural Collapse (NC) \citep{papyan2020prevalence}, where features align with the corresponding classifier weight, inducing overconfidence.
From this view, \textit{periphery} features that are farther from the class center, i.e., those with lower confidence, tend to show prediction changes in retraining.

\noindent \textbf{Correctness Change Analysis.}  \enspace
We evaluated the correctness of each sample in \(\mathcal{D}_f\) under \(\mathcal{M}_r\), then sorted the samples in ascending order based on confidence from \(\mathcal{M}_o.\) 
We applied Kernel Density Estimation to visualize the distributions of correctly and incorrectly predicted samples.
Note that, due to the interpolation regime \citep{gamba2023deepdoubledescentsmooth, block2025machineunlearningoverparameterization}, \(\mathcal{D}_f\) achieved near-perfect accuracy in \(\mathcal{M}_o\). Therefore, we interpret the incorrect samples in \(\mathcal{D}_f\) as those whose labels changed from correct to incorrect.
As shown in \Cref{fig:correctness}, samples with low confidence in \(\mathcal{M}_o\) tended to be misclassified by \(\mathcal{M}_r\), whereas correctly predicted samples were distributed evenly.

\noindent \textbf{Feature Change Analysis.}  \enspace
For qualitative analysis, we extracted features from the penultimate layer for \(\mathcal{M}_{o}\) and \(\mathcal{M}_{r}\).  
We performed T-SNE on samples split by the lower and upper 50\% confidence from \(\mathcal{M}_{o}\).
The results in \Cref{fig:tsne} show that samples with high confidence form tight clusters around the class weight, while low-confidence samples are more peripheral. 
Moreover, in $\mathcal{M}_r$, as low-confidence samples become more intermixed near class boundaries than high-confidence samples.
These results provide visual evidence that low-confidence \textit{periphery} samples undergo greater changes in features and predictions at retraining.

\noindent \textbf{Brief Discussion.} \enspace We identified low-confident \textit{peripheral} samples as the primary drivers of change during retraining and thus interpretable as `` forgotten.''
This trend was consistent across setups. See Appendix \textit{Discrepancy Analysis}.

\section{Method}

\begin{figure*}[t]
	\centering
	\begin{subfigure}[b]{0.32\textwidth}
		\centering
		\includegraphics[width=\linewidth]{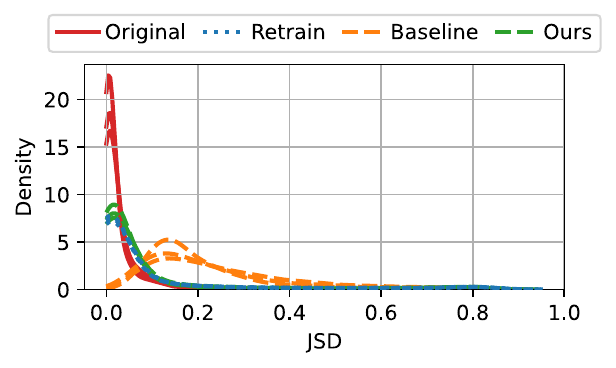}
		\caption{Forget Set}
	\end{subfigure}
	\begin{subfigure}[b]{0.32\textwidth}
		\centering
		\includegraphics[width=\linewidth]{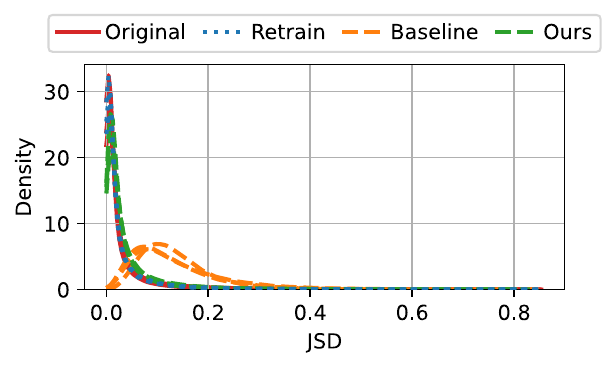}
		\caption{Retain Set}
	\end{subfigure}
	\begin{subfigure}[b]{0.32\textwidth}
		\centering
		\includegraphics[width=\linewidth]{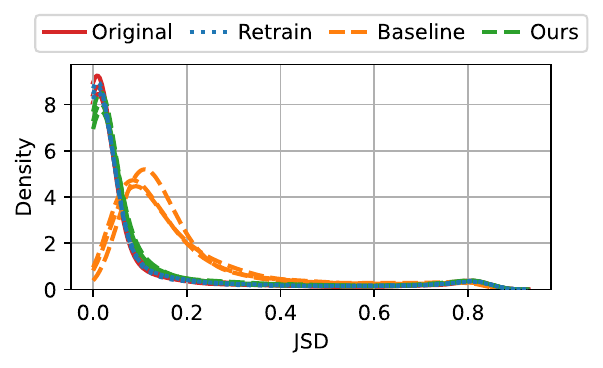}
		\caption{Test Set}
	\end{subfigure}
	\caption{Jenson-Shannon distance distribution between one-hot label vectors and softmax outputs in CIFAR10 / ResNet18 (Setup 1). Our PeriUn alleviates catastrophic forgetting and better approximates the retrain model compared to the SU baseline. We visualize the distribution using the first unlearn seed and all three init seeds, overlaid in the same color.}
    \label{fig:damagelessthreecol}
\end{figure*}

\begin{figure*}[t]
  \centering

  \begin{subfigure}{0.23\textwidth}
    \centering
    \includegraphics[width=\linewidth]{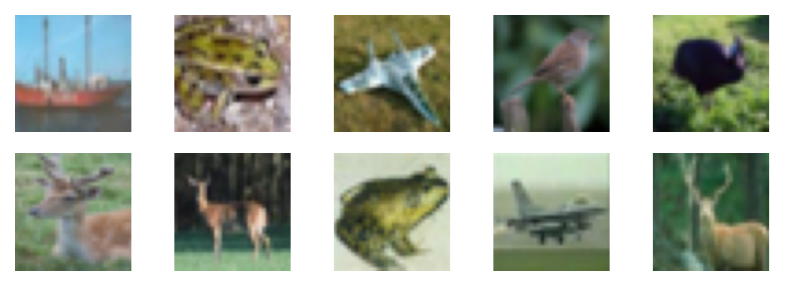}
  \end{subfigure}
  \hfill
  \begin{subfigure}{0.23\textwidth}
    \centering
    \includegraphics[width=\linewidth]{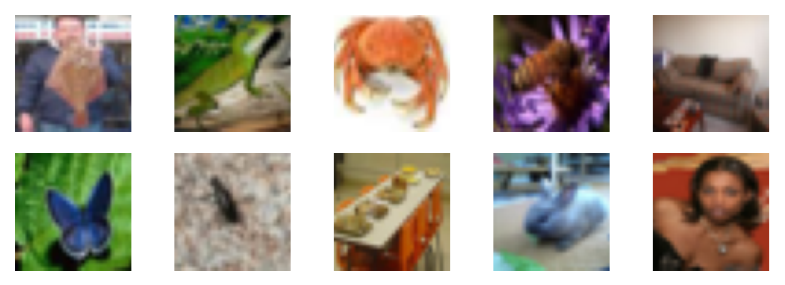}
  \end{subfigure}
  \hfill
  \begin{subfigure}{0.23\textwidth}
    \centering
    \includegraphics[width=\linewidth]{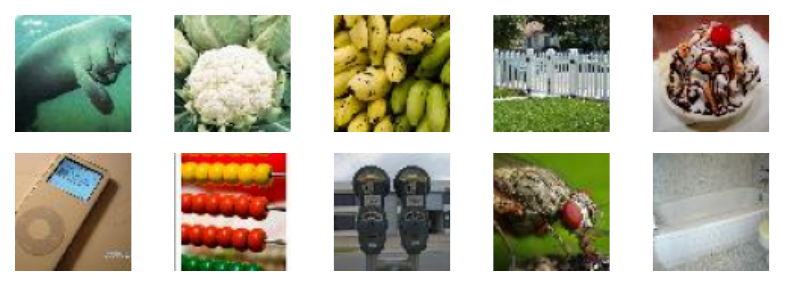}
  \end{subfigure}
  \hfill
  \begin{subfigure}{0.23\textwidth}
    \centering
    \includegraphics[width=\linewidth]{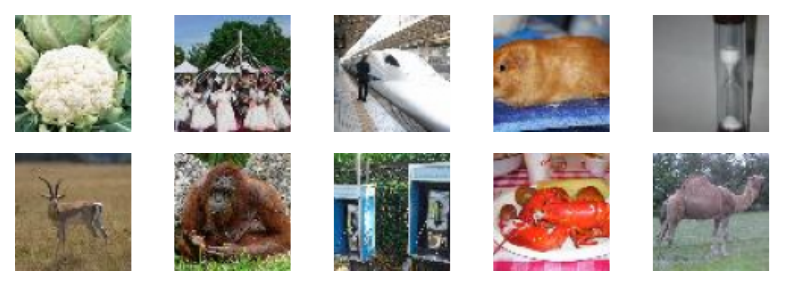}
  \end{subfigure}

  \begin{subfigure}{0.23\textwidth}
    \centering
    \includegraphics[width=\linewidth]{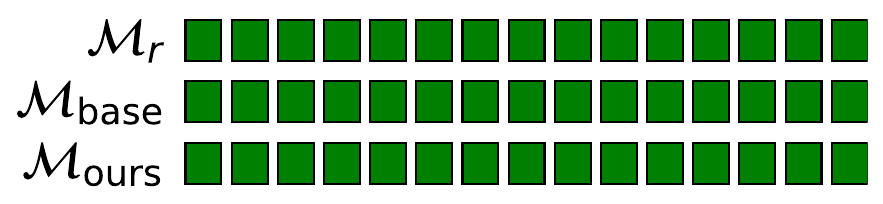}
    \caption{CIFAR10 / ResNet18}
  \end{subfigure}
  \hfill
  \begin{subfigure}{0.23\textwidth}
    \centering
    \includegraphics[width=\linewidth]{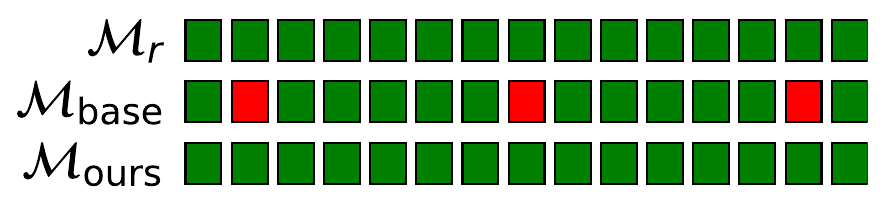}
    \caption{CIFAR100 / ResNet50}
  \end{subfigure}
  \hfill
  \begin{subfigure}{0.23\textwidth}
    \centering
    \includegraphics[width=\linewidth]{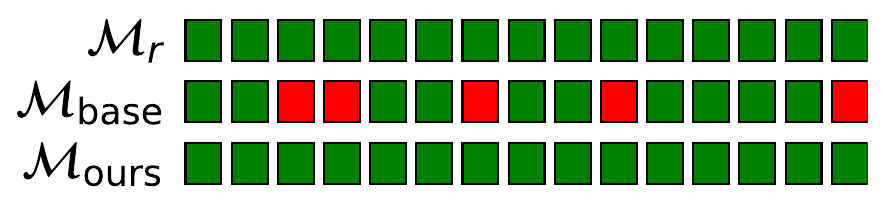}
    \caption{TinyImageNet / ResNet18}
  \end{subfigure}
  \hfill
  \begin{subfigure}{0.23\textwidth}
    \centering
    \includegraphics[width=\linewidth]{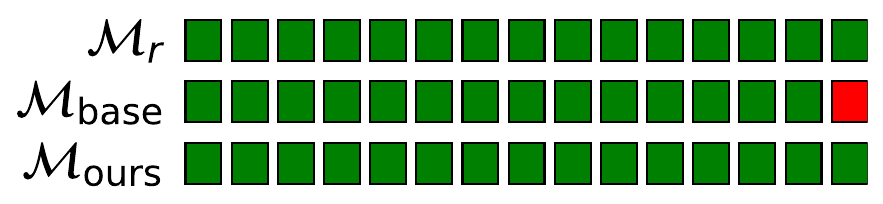}
    \caption{TinyImageNet / VGG16-bn}
  \end{subfigure}

  \caption{Qualitative comparison between PeriUn and Baseline (SU). (Above) Top 10 most confident samples in $\mathcal{M}_f$ selected, mostly visually normal. Images are ordered from left to right, top to bottom, in descending order of confidence. (Below) The top 15 highest-confidence samples are highlighted in red for those that were originally correct but later changed to incorrect (i.e., deletion) by each model. Retrain and Periun tend to preserve high-confidence samples, while Baseline deletes all requested samples, including high-confidence ones that do not need to be removed, which suggests catastrophic forgetting.
}
  \label{fig:random_saliency_4x2}
\end{figure*}

Building on our analysis that \textit{periphery} samples contribute significantly to changes caused by retraining, 
we propose a simple principle to improve approximate unlearning: \emph{unlearn only uncertain}.
We introduce \textbf{Peri}pheral \textbf{Un}learning, \textbf{PeriUn} (see \Cref{fig:main_method}) that selects the subset of the forget set $\mathcal{D}_f$ consisting of low-confidence samples near the decision boundary.
We hypothesize that our principle prevents catastrophic forgetting during the unlearning by safely ignoring confident samples of forget set $\mathcal{D}_f$, which remain unchanged in retraining, and better approximates $\mathcal{M}_r$.

PeriUn is described in \Cref{alg:main}. Given an original model \(\mathcal{M}_o\), a forget set \(\mathcal{D}_{f} = (X_{f}, Y_{f})\), and a selection ratio \(\alpha \in (0,1)\),  
PeriUn constructs a new forget set by selecting the least confident samples in \(\mathcal{D}_{f}\).  
Specifically, for each input \(x_i \in X_f\), the model outputs a probability vector $p_i \leftarrow \mathcal{M}_{o}(x_i)$, and the confidence score is computed as \(c_i \leftarrow \max_j p_{ij}\).
We then compute the threshold \(\tau\) as the \(\alpha\)-quantile of these confidence scores.  
Samples with confidence \(c_i \leq \tau\), i.e., confidence lower than or equal to this threshold \(\tau\), are selected to form the new forget set \(\mathcal{D}^{\text{new}}_{f}\).

\begin{algorithm}[h]
\caption{Selecting low confident samples in \(\mathcal{D}_{f}\)}
\label{alg:main}
\textbf{Input 1}: Original model \(\mathcal{M}_{o}\) \\
\textbf{Input 2}: Forget set \(\mathcal{D}_{f} = (X_{f}, Y_{f})\)\\
\textbf{Parameter}: Number of classes \(K\), Selection ratio \(\alpha\)\\
\textbf{Output}: New forget set \(\mathcal{D}^{new}_{f}\)
\begin{algorithmic}[1]
\STATE Compute probabilities \[p \gets \mathcal{M}_{o}(X_{f}) \in \mathbb{R}^{|\mathcal{D}_{f}| \times K}.\]
\STATE Let confidence $c \in \mathbb{R}^{\lvert \mathcal{D}_{f} \rvert}$ where $c_i \gets \max_j p_{ij}$
\STATE Determine threshold \(\tau \gets \text{Quantile}_\alpha(c)\).
\STATE Select indices \(I = \{ i \mid c_i \leq \tau \}\).
\STATE Extract truncated forget set \(\mathcal{D}^{new}_{f} = \mathcal{D}_{f}[I]\).
\STATE \textbf{return} \(\mathcal{D}^{new}_{f}\)
\end{algorithmic}
\end{algorithm}

\section{Experiments}
Our experiments aim to assess whether PeriUn effectively mimics retraining behavior and improves unlearning performance, thereby addressing our second research question. 
To this end, we pursue the following objectives.

\begin{itemize}
    \item Analyze whether PeriUn produces softmax predictions that more closely resemble those of the retrained model.
    \item Evaluate whether PeriUn enhances existing methods in terms of generalization and privacy performance.
    \item Investigate the impact of the selection ratio $\alpha$.
    \item Assess the compatibility of PeriUn on various baselines.
\end{itemize}

\noindent \textbf{Hyperparameter.}  \enspace
We share hyperparameter search procedures and values in Appendix Hyperparameter Information.

\begin{table*}[t]
	\centering
	{\fontsize{9}{11}\selectfont
		\begin{tabular}{l|cccc|cccc}
			\hline
			  Setup         & \multicolumn{4}{c}{1 - CIFAR10 / ResNet18}      & \multicolumn{4}{c}{2 - CIFAR100 / ResNet50}                                                                                                                                      \\
			           & Forget                                                        & Retain                                                        & Test           & ToW                        & Forget          & Retain          & Test            & ToW                        \\
			Retrain    & 92.55$\pm$0.43                                                & 99.55$\pm$0.11                                                & 92.13$\pm$0.31 &                            & 76.15$\pm$0.98  & 99.97$\pm$0.00  & 76.01$\pm$0.67  &                            \\
			FT         & 95.93$\pm$0.51                                                & 98.51$\pm$0.37                                                & 90.62$\pm$0.49 & 94.18$\pm$0.84             & 74.79$\pm$1.93  & 87.67$\pm$1.41  & 64.86$\pm$1.06  & 76.37$\pm$3.05             \\
			FT+L1      & 95.06$\pm$0.54                                                & 97.90$\pm$0.25                                                & 90.16$\pm$0.41 & 94.00$\pm$0.63             & 99.98$\pm$0.02  & 99.97$\pm$0.00  & 76.18$\pm$0.29  & 75.59$\pm$0.99             \\
			GA         & 99.47$\pm$0.17                                                & 99.46$\pm$0.10                                                & 92.21$\pm$0.29 & 92.60$\pm$0.68             & 99.92$\pm$0.05  & 99.97$\pm$0.00  & 76.19$\pm$0.34  & 75.69$\pm$0.96             \\
			GAGD       & 94.62$\pm$0.58                                                & 99.40$\pm$0.07                                                & 91.52$\pm$0.27 & \underline{97.12$\pm$0.79} & 79.32$\pm$2.60  & 99.04$\pm$0.17  & 70.73$\pm$0.44  & 90.73$\pm$1.67             \\
			RL         & 80.64$\pm$2.01                                                & 98.21$\pm$0.38                                                & 89.70$\pm$0.66 & 84.82$\pm$2.87             & 48.70$\pm$1.20  & 91.88$\pm$1.24  & 64.95$\pm$0.89  & 59.33$\pm$2.40             \\
			SU      & 94.36$\pm$2.12                                                & 99.42$\pm$0.17                                                & 91.46$\pm$0.56 & 96.77$\pm$1.63             & 73.20$\pm$3.13  & 99.95$\pm$0.01  & 73.90$\pm$0.40  & \underline{94.58$\pm$2.49} \\
			SU+ours & 93.92$\pm$0.70                                                & 99.60$\pm$0.10                                                & 91.60$\pm$0.31 & \textbf{97.92$\pm$0.83}    & 76.22$\pm$0.74  & 99.94$\pm$0.01  & 74.62$\pm$0.29  & \textbf{97.78$\pm$0.89}    \\ \hline
		Setup	           & \multicolumn{4}{c}{3 - TinyImagenet / ResNet18} & \multicolumn{4}{c}{4 - TinyImagenet / VGG16-bn}                                                                                                                                  \\
			           & Forget                                                        & Retain                                                        & Test           & ToW                        & Forget          & Retain          & Test            & ToW                        \\
			Retrain    & 64.21$\pm$1.14                                                & 99.98$\pm$0.00                                                & 64.49$\pm$0.34 &                            & 59.90$\pm$1.27  & 99.98$\pm$0.00  & 60.28$\pm$0.39  &                            \\
			FT         & 81.09$\pm$1.52                                                & 96.57$\pm$0.75                                                & 58.00$\pm$0.96 & 75.06$\pm$0.88             & 71.67$\pm$2.21  & 82.66$\pm$2.30  & 49.04$\pm$1.57  & 64.70$\pm$1.38             \\
			FT+L1      & 68.79$\pm$0.90                                                & 90.21$\pm$1.59                                                & 55.79$\pm$0.79 & 78.61$\pm$1.41             & 68.20$\pm$1.08  & 85.96$\pm$1.52  & 50.98$\pm$0.69  & 71.49$\pm$1.03             \\
			GA         & 99.88$\pm$0.07                                                & 99.98$\pm$0.01                                                & 64.48$\pm$0.24 & 64.14$\pm$1.12             & 97.58$\pm$3.54  & 98.58$\pm$3.14  & 57.83$\pm$2.59  & 59.79$\pm$1.21             \\
			GAGD       & 60.22$\pm$2.14                                                & 99.44$\pm$1.07                                                & 60.26$\pm$0.65 & \underline{91.45$\pm$1.94} & 57.74$\pm$16.81 & 87.94$\pm$24.81 & 45.75$\pm$11.96 & 73.48$\pm$25.23            \\
			RL         & 31.99$\pm$1.32                                                & 87.06$\pm$1.11                                                & 52.86$\pm$0.80 & 52.18$\pm$1.93             & 29.63$\pm$1.28  & 86.19$\pm$1.69  & 49.25$\pm$1.12  & 53.51$\pm$2.52             \\
			SU      & 55.41$\pm$1.23                                                & 99.97$\pm$0.01                                                & 62.51$\pm$0.43 & 89.39$\pm$1.53             & 35.64$\pm$1.96  & 99.97$\pm$0.00  & 58.00$\pm$0.32  & \underline{74.01$\pm$1.64} \\
			SU+ours & 64.08$\pm$0.86                                                & 99.97$\pm$0.01                                                & 63.27$\pm$0.27 & \textbf{97.55$\pm$0.68}    & 62.94$\pm$0.80  & 99.98$\pm$0.00  & 59.34$\pm$0.30  & \textbf{96.05$\pm$1.26}    \\ \hline
		\end{tabular}
	}
	\caption{Comparison with existing methods. Forget, retain, and test accuracies, along with ToW scores, are reported across four setups. Applying our method to SU improves performance in all cases.}
    \label{tab:comparision}
\end{table*}

\noindent \textbf{Understanding the Effect of PeriUn.}  \enspace
We claim that PeriUn better approximates the retrained model than baselines, while preventing catastrophic forgetting that would overly alter the output. %
To verify this, we use RL and SU, which directly manipulate features and integrate well with \textbf{PeriUn}, and conduct quantitative and qualitative analyses.

To quantitatively analyze the prediction behavior, we measure the Jensen-Shannon distance (JSD) distributions between one-hot labels and the softmax outputs from \(\mathcal{D}_f\), \(\mathcal{D}_r\), and \(\mathcal{D}_t\) to compare \(\mathcal{M}_o\), \(\mathcal{M}_r\), baseline, and baseline+PeriUn.
\Cref{fig:damagelessthreecol} presents the results for CIFAR10 and ResNet18 (Setup 1) on SU.
The results indicate that baseline+PeriUn yields output distributions consistently closer to $\mathcal{M}_r$ than the baseline.
Moreover, the baseline exhibits higher JSD after unlearning, reflecting lower confidence in the correct class and weaker convergence toward Neural Collapse, which suggests more severe catastrophic forgetting.
These suggest that, unlike baseline methods that indiscriminately modify all forget set samples and cause redundant damage, our method selectively unlearns, more closely mimicking retraining without catastrophic forgetting.

For qualitative analysis, we confirmed in \Cref{fig:random_saliency_4x2} that PeriUn imitates the correctness behavior of the retrained model, which preserves high-confidence ones.
Visual inspection showed the top 10 highest-confidence \(\mathcal{D}_f\) samples of \(\mathcal{M}_o\) are regular and recognizable images likely classifiable and indeed remain correct without inclusion in retraining.
Furthermore, an analysis of the top 15 high-confidence samples that changed from correct to incorrect predictions reveals that both the retrained model and PeriUn successfully preserved them, unlike the baseline. This indicates that the baseline removes unnecessary samples to be erased according to retraining, whereas PeriUn effectively replicates the behavior of retrained model. Additional results for this section are in the appendix under \emph{Comparison with Baseline}.

\noindent \textbf{Comparison with the State-of-the-art Baselines.}  \enspace
As shown in \Cref{tab:comparision}, we observed consistent improvements in ToW performance when applying PeriUn to SU, the state-of-the-art unlearning approach based on Random Label.
Our approach, SU$+$PeriUn, outperforms the best ToW scores of all existing baselines across all experimental setups.
Moreover, compared to the best-performing baseline in each case, our method achieves average ToW improvements of 0.8 points on CIFAR-10 with ResNet18, 3.2 points on CIFAR-100 with ResNet50, 6.1 points on TinyImageNet with ResNet18, and 22.04 points on TinyImageNet with VGG16-bn.
These results indicate that PeriUn achieves an ToW score of 96 across all setups, demonstrating its ability to mimic the performance of the retrained model closely.

\begin{table}[h]
	\centering
	{\fontsize{9}{11}\selectfont
		\begin{tabular}{cc|cc}
			\hline
			\multicolumn{2}{c|}{Setup} & 1 - C10 / Res18      & 2 - C100 / Res50                                                           \\ \hline
			\multirow{5}{*}{$\alpha$}    & 0.1          & \textbf{97.92$\pm$0.83 (+1.14)} & 84.14$\pm$0.59 (-10.44)          \\
			                             & 0.3          & 97.56$\pm$1.69 (+0.79)          & \textbf{97.78$\pm$0.89 (+3.20)}  \\
			                             & 0.5          & 97.06$\pm$1.31 (+0.29)          & 97.12$\pm$1.48 (+2.54)           \\
			                             & 0.7          & 96.95$\pm$1.74 (+0.18)          & 94.09$\pm$3.22 (-0.48)           \\
			                             & 1.0          & 96.77$\pm$1.63 (Base)           & 94.58$\pm$2.49 (Base)            \\ \hline
			\multicolumn{2}{c|}{Setup} & 3 - TinyIn / Res18 & 4 - TinyIn / VGG16                                                       \\ \hline
			\multirow{5}{*}{$\alpha$}    & 0.1          & 80.59$\pm$0.90 (-8.80)          & 76.62$\pm$1.09 (+2.62)           \\
			                             & 0.3          & 89.12$\pm$1.01 (-0.27)          & 89.00$\pm$1.40 (+15.00)          \\
			                             & 0.5          & \textbf{97.55$\pm$0.68 (+8.16)} & \textbf{96.05$\pm$1.26 (+22.04)} \\
			                             & 0.7          & 96.41$\pm$1.20 (+7.01)          & 89.17$\pm$1.30 (+15.17)          \\
			                             & 1.0          & 89.39$\pm$1.53 (Base)           & 74.01$\pm$1.64 (Base)            \\ \hline
		\end{tabular}
	}
	\caption{ToW measure across different selection ratios. As the ratio decreases from 1.0, performance initially increases and then decreases, exhibiting an inverted U-shaped trend.}
    \label{tab:alpha_abla}
\end{table}

\noindent \textbf{Effect of Selection Ratio.}  \enspace
We conduct an ablation study on the selection ratio $\alpha$.
Since high-confidence samples often remain correctly predicted after retraining, excluding them may preserve key information and improve similarity to $\mathcal{M}_r$.
Conversely, excluding too many samples can hinder unlearning.
We thus expect an inverted U-shaped trend with respect to the selection ratio \(\alpha\), where performance improves up to a certain point and then declines.
This supports the notion that our periphery-based selection is an adjustable algorithm that corresponds to our hypothesis.

To test this, under the SU$+$PeriUn setting, we varied the selection ratio $\alpha \in \{0.1, 0.3, 0.5, 0.7, 1.0\}$ and measured ToW as in \Cref{tab:alpha_abla}.
Across all setups, we observed an inverted U-shape trend in performance with respect to the selection ratio.
It validates that excluding high-confidence samples up to a certain level can contribute positively to performance.
This provides empirical support for the design rationale of PeriUn. 
While a slight performance drop was observed at $\alpha = 0.7$ in the CIFAR100 / ResNet50 (Setup 2), it was not statistically significant (paired $t$-test, $p = 0.35$).

\begin{table}[t]
\centering
{\fontsize{9}{11}\selectfont
\begin{tabular}{l|cc|cc} \hline
          \multicolumn{1}{c|}{Setup} & \multicolumn{2}{c}{1 - C10 / Res18}      & \multicolumn{2}{c}{2 - C100 / Res50 }  \\ \hline
           & \multicolumn{1}{c}{MIA}      & \multicolumn{1}{c}{Gap}    & \multicolumn{1}{c}{MIA}     & \multicolumn{1}{c}{Gap}     \\
Retrain    & 14.75$\pm$0.61               & \multicolumn{1}{l}{}           & 46.07$\pm$1.23              & \multicolumn{1}{l}{}            \\
FT         & 9.73$\pm$1.02                & 5.02                           & 30.99$\pm$1.81              & 15.09                           \\
FT+L1      & 10.88$\pm$0.80               & \underline{ 3.87}                     & 1.53$\pm$0.21               & 44.54                           \\
GAGD       & 9.50$\pm$0.83                & 5.25                           & 35.45$\pm$3.76              & \textbf{10.62}                  \\
GA         & 3.74$\pm$0.74                & 11.01                          & 1.76$\pm$0.19               & 44.31                           \\
RL         & 47.62$\pm$4.42               & 32.87                          & 73.69$\pm$1.29              & 27.61                           \\
SU      & 23.28$\pm$6.07               & 8.53                           & 91.79$\pm$1.09              & 45.71                           \\
SU+ours & 12.02$\pm$1.23               & \textbf{2.73}                  & 33.84$\pm$0.40              & \underline{ 12.24}                     \\ \hline
           \multicolumn{1}{c|}{Setup} & \multicolumn{2}{c}{3 - TinyIn / Res18}      & \multicolumn{2}{c}{4 - TinyIn / VGG16}     \\ \hline
           & \multicolumn{1}{c}{MIA}      & \multicolumn{1}{c}{Gap}    & \multicolumn{1}{c}{MIA}     & \multicolumn{1}{c}{Gap}     \\
Retrain    & 63.24$\pm$1.10               & \multicolumn{1}{l}{}           & 61.37$\pm$1.14              & \multicolumn{1}{l}{}            \\
FT         & 31.31$\pm$1.56               & 31.93                          & 29.04$\pm$1.58              & 32.33                           \\
FT+L1      & 37.74$\pm$0.99               & 25.50                          & 33.73$\pm$0.99              & 27.64                           \\
GAGD       & 56.44$\pm$3.67               & \textbf{6.79}                  & 48.56$\pm$11.95             & \underline{ 12.81}                     \\
GA         & 4.71$\pm$0.47                & 58.53                          & 6.84$\pm$4.26               & 54.53                           \\
RL         & 76.84$\pm$1.50               & 13.60                          & 78.03$\pm$1.28              & 16.66                           \\
SU      & 96.42$\pm$0.45               & 33.18                          & 81.73$\pm$0.96              & 20.36                           \\
SU+ours & 50.90$\pm$0.30               & \underline{ 12.34}                    & 49.75$\pm$0.14              & \textbf{11.62}  \\ \hline                
\end{tabular}
}
\caption{Confidence-based MIA gap. Our SU$+$PeriUn achieves comparable or superior performance to existing methods and significantly improves the MIA gap of SU.}
\label{tab:mia_conf}
\end{table}

\noindent \textbf{Method Evaluation Based on MIA Gap.}  \enspace
To evaluate whether our method not only improves ToW but also better approximates the privacy behavior of the retrained model, we measure MIA accuracy for each method and compute the gap to the retrained model, quantifying how closely each method mimics privacy behavior.
Following \citet{zhao2024makesunlearninghard, jia2024modelsparsitysimplifymachine}, we primarily focus on confidence-based MIA. Additionally, we evaluate MIA performance using two further metrics, output probability and entropy, to provide a more comprehensive analysis. Results for these additional metrics are reported in the appendix \emph{MIA Results}.
As shown in \Cref{tab:mia_conf}, SU$+$PeriUn significantly reduces the confidence-based MIA gap compared to SU, suggesting improved alignment with the privacy characteristics of the retrain model.
This highlights that SU in itself is vulnerable in privacy despite its strong ToW performance.
In contrast, our method enhances both utility and privacy, demonstrating better alignment with the retrain model.
Notably, SU$+$PeriUn achieves the best MIA gap in CIFAR10-ResNet18 and TinyImagenet-ResNet18 (Setups 1 and 3), and ranks second-best in the remaining cases.
These results indicate that our approach provides a favorable trade-off between utility and privacy guarantees in unlearning.

\begin{table}[t]
	\centering
	{\fontsize{9}{11}\selectfont
		\begin{tabular}{l|ll} \hline
			\multicolumn{1}{c|}{Setup} & \multicolumn{1}{c}{1 - C10 / Res18}      & \multicolumn{1}{c}{2 - C100 / Res50 }     \\ \hline
			GAGD                         & 97.12$\pm$0.79                   & 90.73$\pm$1.67                   \\
			GAGD+ours                    & 96.64$\pm$0.75 (-0.48)           & \textbf{93.00$\pm$1.55 (+2.27)}  \\
			GA                           & 92.60$\pm$0.68                   & 75.69$\pm$0.96                   \\
			GA+ours                      & 92.77$\pm$0.65 (+0.17)           & 75.73$\pm$0.89 (+0.04)           \\
			RL                           & 84.82$\pm$2.87                   & 59.33$\pm$2.40                   \\
			RL+ours                      & \textbf{97.76$\pm$0.43 (+12.94)}  & \textbf{95.65$\pm$1.46 (+36.32)} \\ \hline
			\multicolumn{1}{c|}{Setup} & \multicolumn{1}{c}{3 - TinyIn / Res18} & \multicolumn{1}{c}{4 - TinyIn / VGG16} \\ \hline
			GAGD                         & 91.45$\pm$1.94                   & 73.48$\pm$25.23                  \\
			GAGD+ours                    & \textbf{93.07$\pm$1.71 (+1.62)}  & \textbf{76.78$\pm$8.83 (+3.30)}  \\
			GA                           & 64.14$\pm$1.12                   & 59.79$\pm$1.21                   \\
			GA+ours                      & 64.11$\pm$1.10 (-0.03)           & 59.91$\pm$1.35 (+0.12)           \\
			RL                           & 52.18$\pm$1.93                   & 53.51$\pm$2.52                   \\
			RL+ours                      & \textbf{86.90$\pm$1.88 (+34.72)} & \textbf{79.99$\pm$1.91 (+26.48)} \\ \hline
		\end{tabular}
	}
    \caption{Our method was applied to GAGD, GA, and RL.
    Notable improvements are observed in RL, the base of SU.} %
    \label{tab:otherpours} 
\end{table}

\noindent \textbf{Applying PeriUn to Other Baselines.}  \enspace
We applied our method to RL, GA, and GAGD models, all of which utilize the forget set $\mathcal{D}_f$.
In \Cref{tab:otherpours}, RL exhibited a significant performance improvement similar to SalUn, with a notable ToW gain of 36.32 points on Setup 2.
GAGD showed performance improvements in most settings.
We attribute the effectiveness of PeriUn on RL and SU, where samples in $\mathcal{D}_f$ are intentionally reassigned to incorrect classes.
By discarding non-peripheral samples, PeriUn reduces over-deletion and helps RL and SU better mimic the behavior of a retraining.
In contrast, GA and GAGD apply negative gradients that perturb the entire feature space. This disruptive nature likely undermines the benefits of PeriUn, as selective removal of peripheral samples becomes less meaningful.

\section{Discussion}
Our work introduces PeriUn, a method motivated by retrained model analysis, memorization, and coreset. Through this approach, we address two key research questions.
First, we observe that the original and retrained models differ on the forget set, and attribute these differences primarily to peripheral samples.
Second, we propose PeriUn, which improves unlearning while reducing unintended damage.

Moreover, our observation that samples altered to incorrect predictions tend to be low-confidence peripheral samples offers a practical connection to memorization.
Prior work has shown a correlation between average confidence across training and memorization \citep{jiang2021char}, and leveraged average confidence as a proxy \citep{zhao2024makesunlearninghard}.
In contrast, we find that the relative ranking of post-training confidence alone is sufficient to identify such vulnerable samples, without requiring per-epoch tracking.
This significantly reduces computational overhead and offers a direct bridge between memorization and unlearning.

Lastly, we show that unlearning a carefully selected subset is sufficient to approximate retraining and even improves performance across all metrics.
The performance of PeriUn validates our hypothesis that a coreset-like subset exists in unlearning and that removing it selectively is beneficial, a notion we explore here for the first time.

\section{Conclusion}
\textbf{PeriUn} presents a novel yet simple unlearning approach that mimics retraining based on empirical analyses.
This work advances unlearning by introducing a selective strategy, which balances privacy and performance more effectively than indiscriminate forgetting.
Our findings highlight a promising direction that leverages feature space properties for effective and selective forgetting.

\section{Acknowledgments}
We are deeply grateful to Inwon Lee for her invaluable support and encouragement. We are also grateful to Won-Seok Choi for insightful scholarly discussions.

This work was partly supported by the IITP (RS-2021-II212068-AIHub/10\%, RS-2021-II211343-GSAI/10\%, RS-2022-II220951-LBA/15\%, RS-2022-II220953-PICA/15\%), NRF (RS-2024-00353991-SPARC/15\%, RS-2023-00274280-HEI/15\%), KEIT (RS-2024-00423940/10\%), and Gwangju Metropolitan City (Artificial intelligence industrial convergence cluster development project/10\%) grant funded by the Korean government.

\bibliography{aaai2026}

\clearpage
\appendix
\section{Definition of MIA gap}
Given model \(\mathcal{M}\), membership inference attack (MIA) introduce a binary classifier using the \(\kappa\)-outputs (e.g., confidence) of \(\mathcal{M}\).
The classifier is trained to distinguish training-exposed (\(\mathcal{D}_r\)) from unexposed (\(\mathcal{D}_t\)) samples.
Denote the success rate of the MIA as \(\operatorname{MIA}(\mathcal{M}, \mathcal{D}_f; \kappa)\).
\begin{definition}[Membership Inference Attack Gap]
The MIA gap between $\mathcal{M}_{r}$ and $\mathcal{M}_{u}$ is then defined as the absolute difference in their MIA success rates:
\begin{equation}
\begin{aligned}
    &\text{MIAgap}(\mathcal{M}_{r}, \mathcal{M}_{u}, \mathcal{D}_{f};\, \kappa) \\
    & = \big| \operatorname{MIA}(\mathcal{M}_{r}, \mathcal{D}_{f};\, \kappa) - \operatorname{MIA}(\mathcal{M}_{u}, \mathcal{D}_{f};\, \kappa) \big|.
\end{aligned}
\end{equation}
\end{definition}
\noindent A smaller MIA gap indicates that $\mathcal{M}_{u}$ closely approximates $\mathcal{M}_{r}$, reflecting similar privacy guarantees.

\section{Further Work}
As supported by our qualitative and quantitative analysis, avoiding unnecessary deletions mitigates catastrophic forgetting.
We hypothesize that the effectiveness of \textbf{PeriUn} in preventing catastrophic forgetting stems from its ability to reduce feature space distortion and limit the propagation of such damage across dispersed regions where forget set samples exist.
Recent studies have shown that the forget and retain sets are often inseparable in the feature space \citep{jeung2025duskunlearnsharedknowledge}, and that high entanglement between these sets makes unlearning particularly challenging \citep{zhao2024makesunlearninghard}.
Based on these findings, indiscriminate removal of samples is likely to affect not only the forget set but also neighboring retain samples, potentially causing widespread degradation across multiple regions of the feature space.
This phenomenon may align with the concept of local elasticity introduced by \citet{he2020localelasticityneuralnetworks}, which suggests that training a neural network influences a neighborhood of samples in the feature space.
Therefore, understanding the connection between unlearning and local elasticity offers a promising avenue for future theoretical exploration.

\section{Additional Unlearning Methods}
\citet{chen2023boundaryunlearning} performed unlearning by encouraging the forget set to be classified as the nearest incorrect class, rather than its original class.
\citet{lin2023erm} proposed an unlearning approach that leverages knowledge transfer techniques to convey only the information from the retain set.
\citet{foster2023fastmachineunlearningretraining} introduced a method based on Fisher Information to identify and remove parameters that contain information about the forget set.
Additionally, \citet{chundawat2023badteachinginduceforgetting} employed a distillation-based approach with a random model to ensure the final model behaves as if it had never learned the forget set.
\citet{cotogni2024duckdistancebasedunlearningcentroid} proposed shifting the features of forget samples toward the centroid of their nearest class as a way to achieve unlearning.
\citet{di2024labelsmoothingimprovesmachine} improved the traditional gradient ascent-based method by incorporating label smoothing for better performance.
Similar to \citet{fan2024salunempoweringmachineunlearning}, methods based on saliency information have also been proposed, applying selective parameter masking \citep{torkzadehmahani2024improvedlocalizedmachineunlearning, malekmohammadi2025sharpnessawareparameterselectionmachine}.

\section{Pseudocode of PeriUn}
\Cref{lst:lowconf-forget} presents the pseudocode of our data selection method implemented in PyTorch \citep{paszke2019pytorchimperativestylehighperformance}.

\section{Hyperparameter Information}

\subsubsection{Original Model and Retrain Model hyperparameters}
The hyperparameters used to train the original and retrained models are listed in \Cref{tab:hyper1} and \Cref{tab:hyper2}. The implementation is based on PyTorch \citep{paszke2019pytorchimperativestylehighperformance}.

\subsubsection{Hyperparamter Searching Method}
As shown in \Cref{tab:hyperparam_searchrange}, we performed hyperparameter search using the same range provided in \citet{zhao2024makesunlearninghard}. The search was conducted using an unlearning of the first seed and an initialization of the first seed.
The selected hyperparameters were then used for all experiments.

\subsubsection{Hyperparameters of Baseline Method}
The optimal values were selected as summarized in \Cref{tab:opt_hyperparam}.
All quantitative and qualitative experiments were conducted using these optimized hyperparameters, ensuring fairness and enabling us to investigate each model architecture under its optimal configuration.

\subsubsection{Hyperparameters of PeriUn$+$[GA, GAGD, RL, SU]}
\Cref{tab:optsu_hyperparam} reports the hyperparameters of PeriUn+SU used for the experiments presented in \textit{Comparison with the State-of-the-art Baselines} and \textit{Effect of Selection Ratio}.
\Cref{tab:gapours}, \Cref{tab:gagdpours}, and \Cref{tab:rlpours} provides the hyperparameters used to reproduce \textit{
Applying PeriUn to Other Baselines}.
In particular, the results for RL in \Cref{tab:rlpours} exhibit an inverted U-shape relationship between $\alpha$ and ToW performance, consistent with the interpretation provided in the main results \textit{Effect of Selection Ratio}, thereby supporting our analysis.

\section{Execution Time Analysis}
\Cref{tab:time} reports the additional computational time required to apply the PeriUn across different setups.
For comparison, we also report the time for a single epoch of SOTA unlearning methods, SU and GAGD.
PeriUn requires less than 5 seconds in all cases before unlearning.
Contrarily, even a single epoch of baseline takes longer, ranging from 10 to over 130 seconds, depending on the setup.
Considering that unlearning requires 5 to 10 epochs, the additional time incurred by PeriUn is negligible.
This demonstrates PeriUn achieves its goal without compromising time efficiency.

\section{Confidence Distribution Results}
\begin{figure*}[htbp]
        \centering
        \begin{subfigure}[b]{\linewidth}
            \centering
            \includegraphics[width=0.9\linewidth]{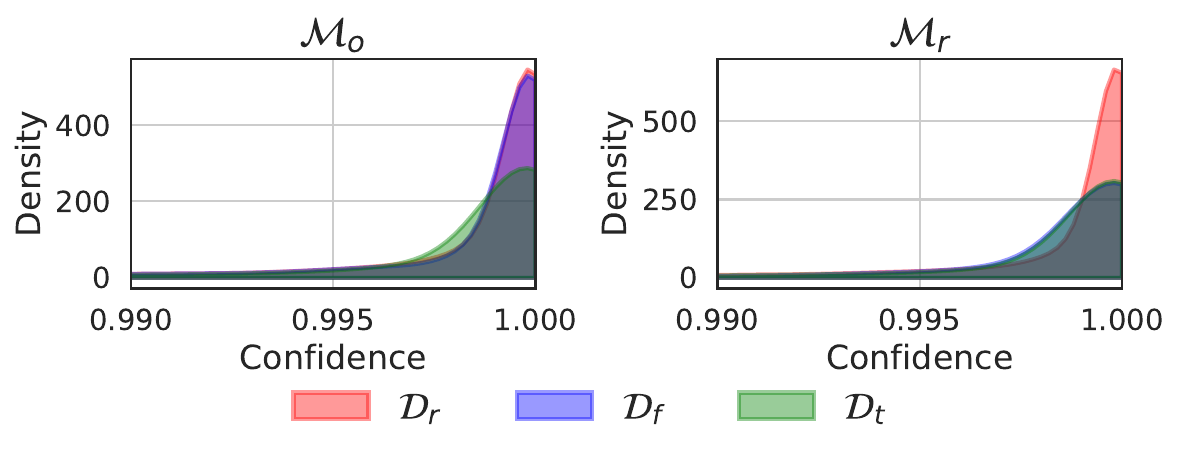}
            \caption{Unlearn seed 0}
        \end{subfigure}
        \begin{subfigure}[b]{\linewidth}
            \centering
            \includegraphics[width=0.9\linewidth]{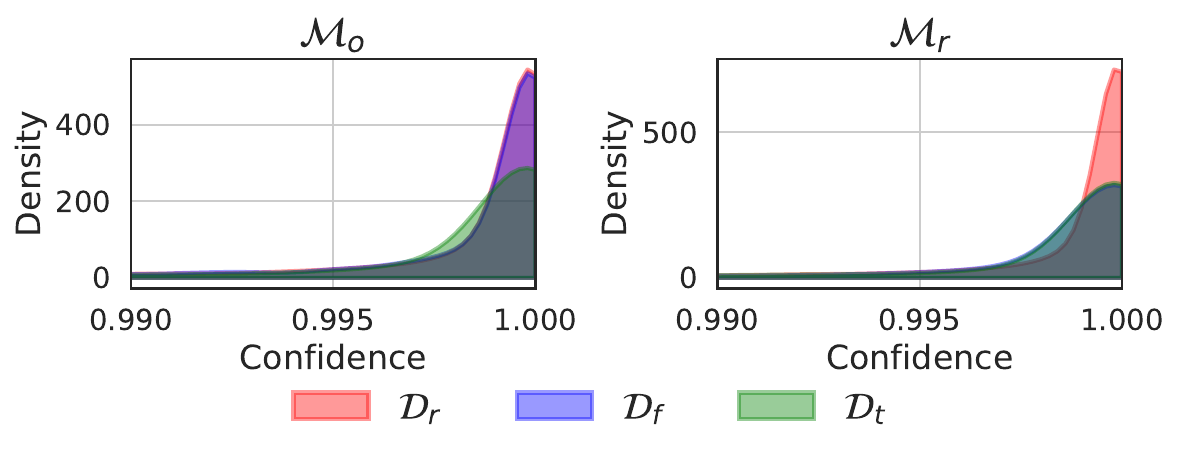}
            \caption{Unlearn seed 1}
        \end{subfigure}
        \begin{subfigure}[b]{\linewidth}
            \centering
            \includegraphics[width=0.9\linewidth]{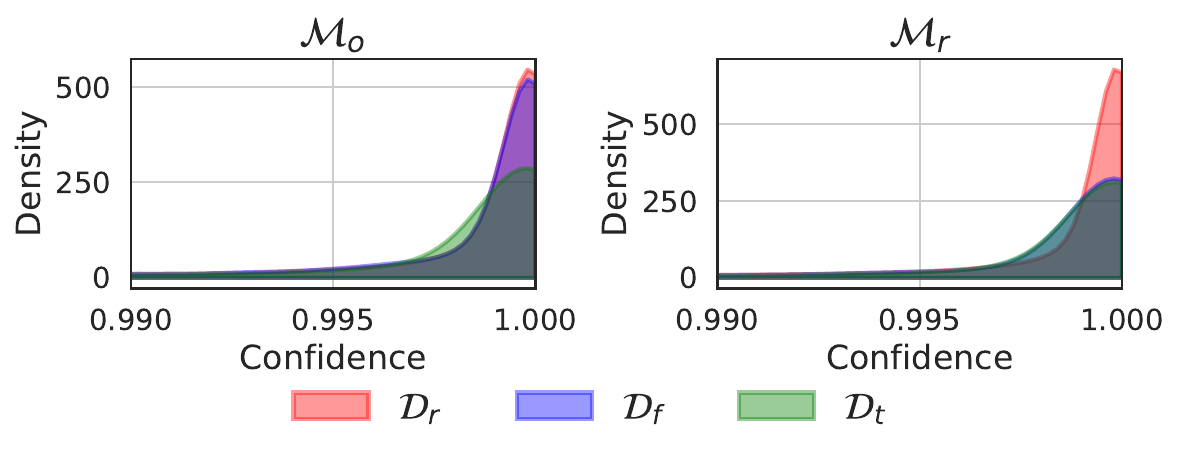}
            \caption{Unlearn seed 2}
        \end{subfigure}
        \caption{CIFAR10-ResNet18 Confidence Distribution}
        \label{fig:conf_dist_appendix_cifar10-resnet18}
    \end{figure*}

\begin{figure*}[htbp]
        \centering
        \begin{subfigure}[b]{\textwidth}
            \centering
            \includegraphics[width=0.9\linewidth]{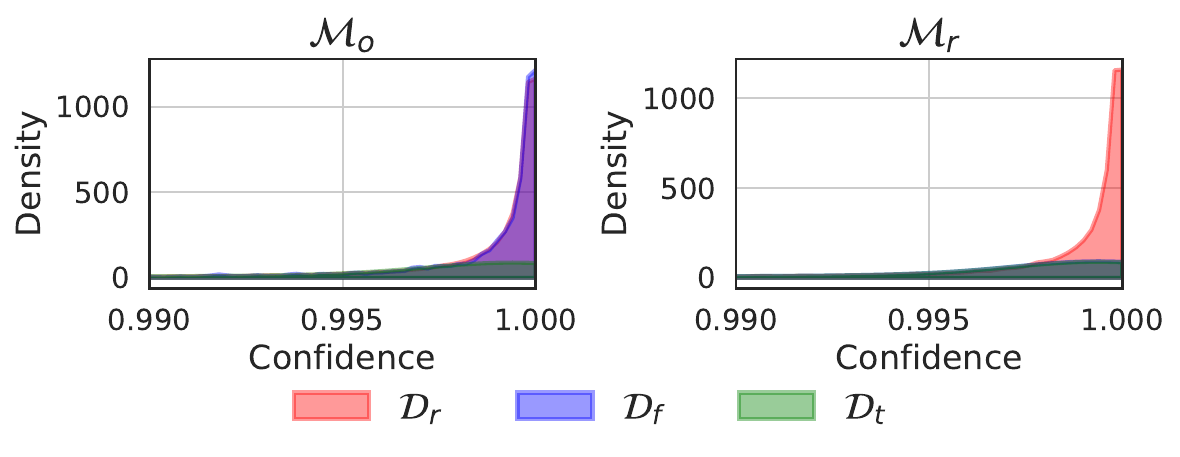}
            \caption{Unlearn seed 0}
        \end{subfigure}
        \begin{subfigure}[b]{\textwidth}
            \centering
            \includegraphics[width=0.9\linewidth]{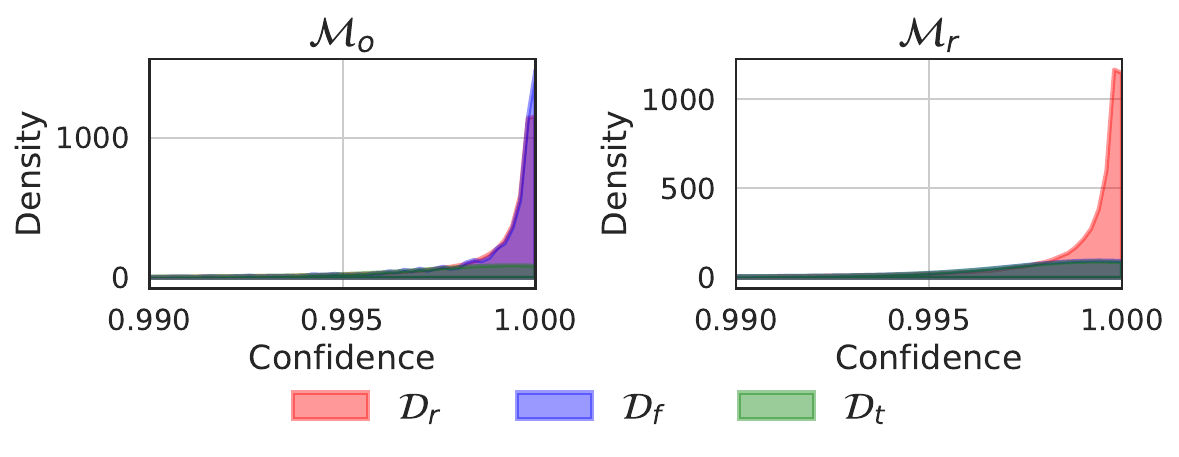}
            \caption{Unlearn seed 1}
        \end{subfigure}
        \begin{subfigure}[b]{\textwidth}
            \centering
            \includegraphics[width=0.9\linewidth]{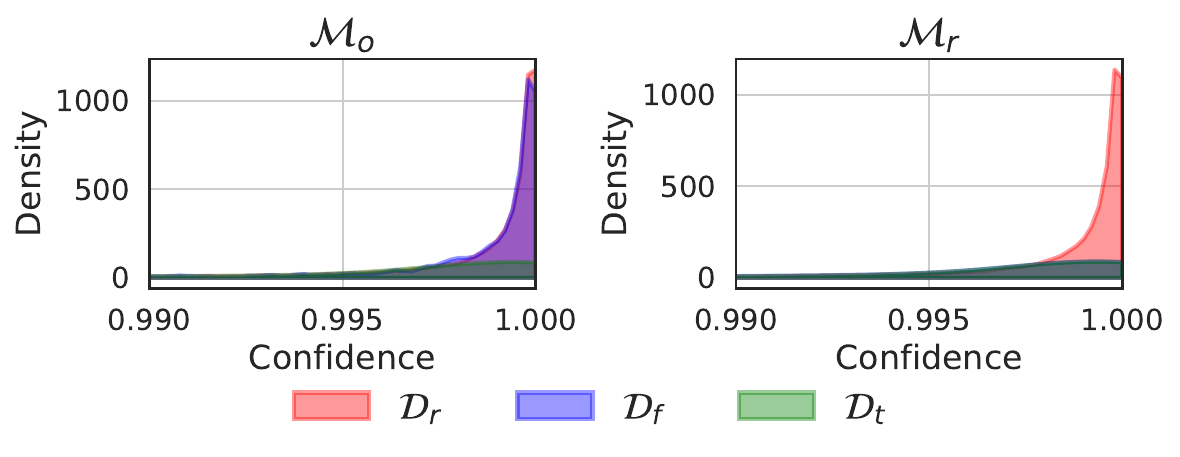}
            \caption{Unlearn seed 2}
        \end{subfigure}
        \caption{CIFAR100-ResNet50 Confidence Distribution}
        \label{fig:conf_dist_appendix_cifar100-resnet50}
    \end{figure*}

\begin{figure*}[htbp]
        \centering
        \begin{subfigure}[b]{\textwidth}
            \centering
            \includegraphics[width=0.9\linewidth]{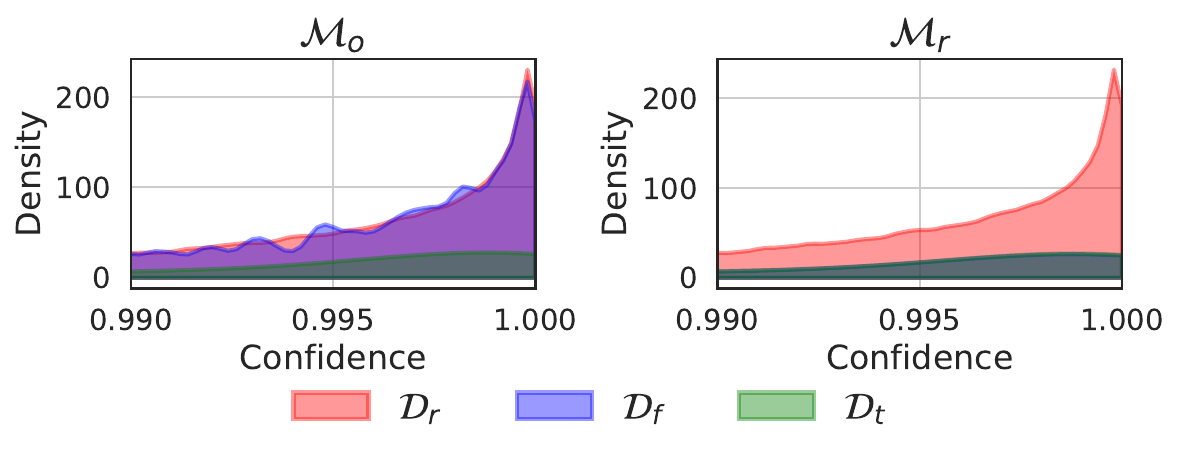}
            \caption{Unlearn seed 0}
        \end{subfigure}
        \begin{subfigure}[b]{\textwidth}
            \centering
            \includegraphics[width=0.9\linewidth]{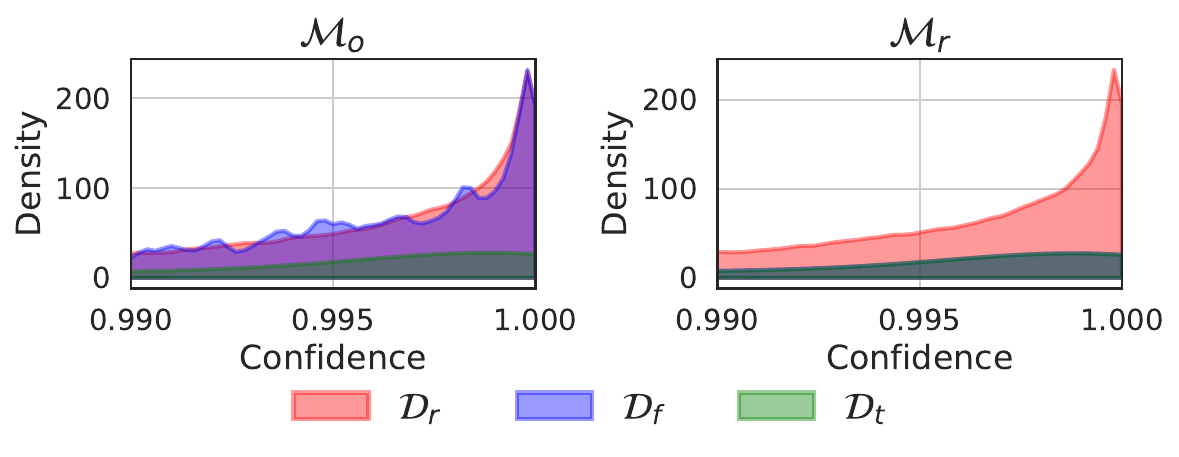}
            \caption{Unlearn seed 1}
        \end{subfigure}
        \begin{subfigure}[b]{\textwidth}
            \centering
            \includegraphics[width=0.9\linewidth]{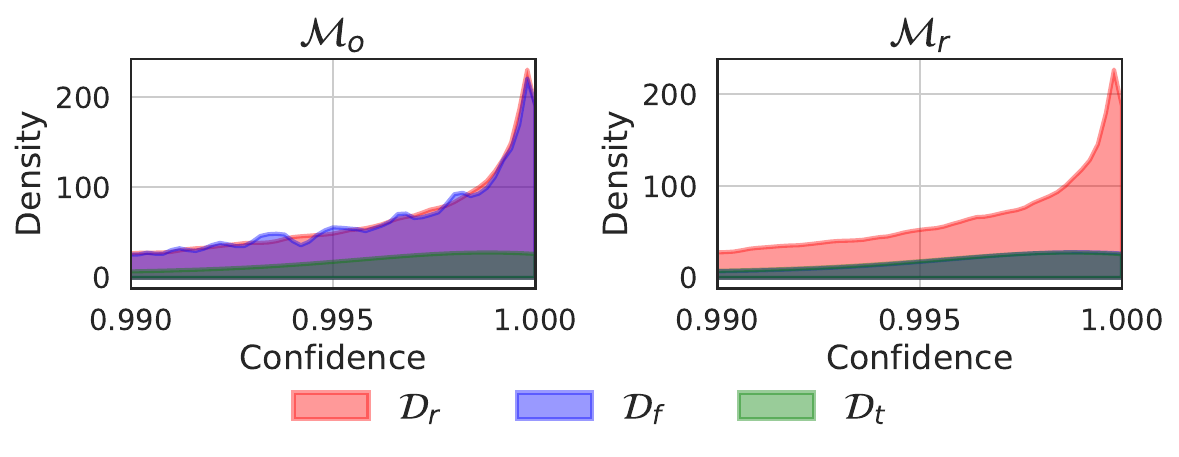}
            \caption{Unlearn seed 2}
        \end{subfigure}
        \caption{TinyImageNet-ResNet18 Confidence Distribution}
        \label{fig:conf_dist_appendix_TinyImagenet-resnet18}
    \end{figure*}

\begin{figure*}[htbp]
        \centering
        \begin{subfigure}[b]{\textwidth}
            \centering
            \includegraphics[width=0.9\linewidth]{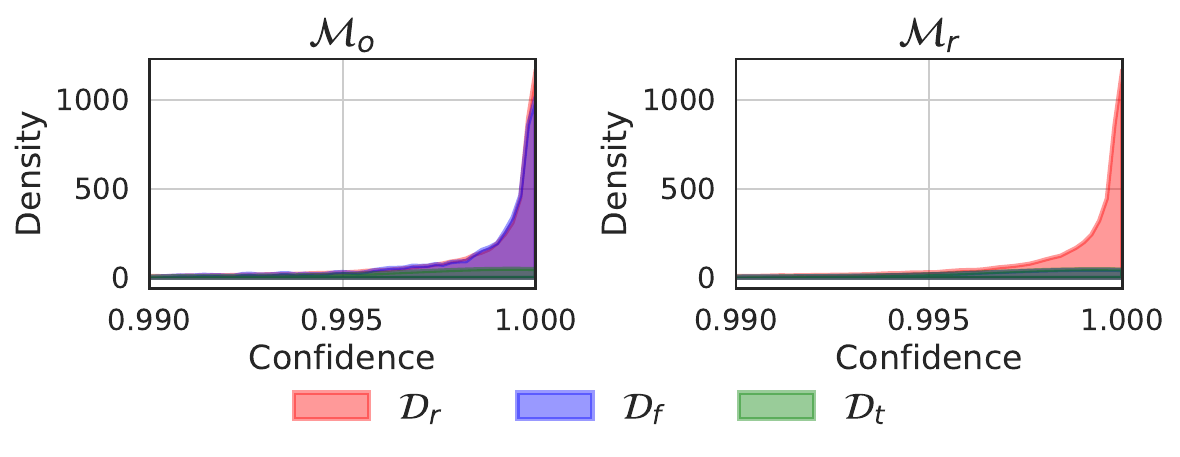}
            \caption{Unlearn seed 0}
        \end{subfigure}
        \begin{subfigure}[b]{\textwidth}
            \centering
            \includegraphics[width=0.9\linewidth]{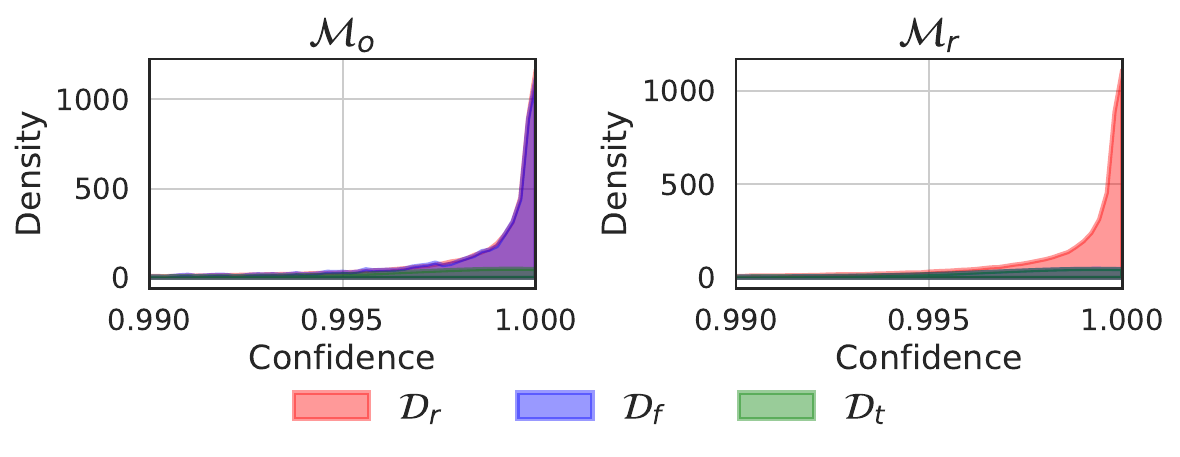}
            \caption{Unlearn seed 1}
        \end{subfigure}
        \begin{subfigure}[b]{\textwidth}
            \centering
            \includegraphics[width=0.9\linewidth]{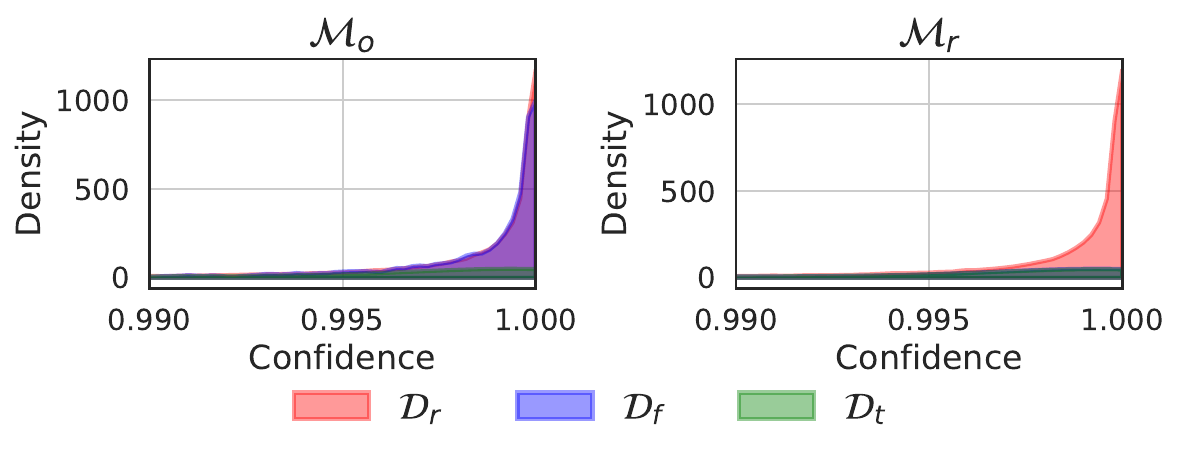}
            \caption{Unlearn seed 2}
        \end{subfigure}
        \caption{TinyImageNet-VGG16-bn Confidence Distribution}
        \label{fig:conf_dist_appendix_TinyImagenet-vgg16_bn}
    \end{figure*}

We provide confidence distribution plots, extending \textit{Observation 1: Seen vs. Unseen} in the main text. Specifically, we present KDE estimates of the confidence scores output by $\mathcal{M}_{o}$ and $\mathcal{M}_{r}$ when evaluated on $\mathcal{D}_{f}$, $\mathcal{D}_{r}$, and $\mathcal{D}_{t}$, using larger figure size and multiple unlearning seeds.
As discussed in the main text, seen data exhibit overconfident distributions, while unseen data show relatively underconfident behavior compared to the seen data.
Refer to \Cref{fig:conf_dist_appendix_cifar10-resnet18}, \Cref{fig:conf_dist_appendix_cifar100-resnet50}, \Cref{fig:conf_dist_appendix_TinyImagenet-resnet18}, and \Cref{fig:conf_dist_appendix_TinyImagenet-vgg16_bn}.

\section{Discrepancy analysis}
\begin{figure}[htbp]
        \centering
        \begin{subfigure}[b]{0.9\columnwidth}
            \centering
            \includegraphics[width=\linewidth]{figures/conf_change/cifar10_resnet18_unlearn_seed_0_confidence_diff}
            \caption{Unlearn seed 0}
        \end{subfigure}
        \begin{subfigure}[b]{0.9\columnwidth}
            \centering
            \includegraphics[width=\linewidth]{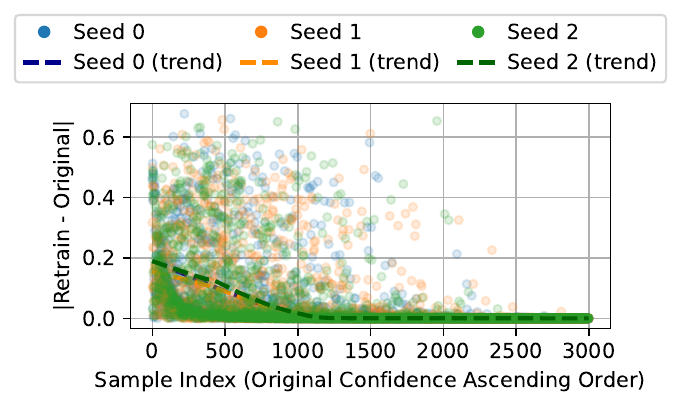}
            \caption{Unlearn seed 1}
        \end{subfigure}
        \begin{subfigure}[b]{0.9\columnwidth}
            \centering
            \includegraphics[width=\linewidth]{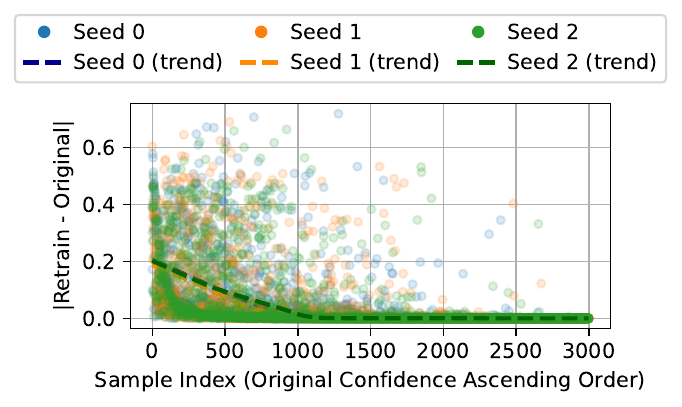}
            \caption{Unlearn seed 2}
        \end{subfigure}
        \caption{CIFAR10-ResNet18 Confidence difference of forget set}
        \label{fig:cifar10resnet18cd}
    \end{figure}

\begin{figure}[htbp]
        \centering
        \begin{subfigure}[b]{0.9\columnwidth}
            \centering
            \includegraphics[width=\linewidth]{figures/conf_change/cifar10_resnet18_unlearn_seed_0_correctness_kde}
            \caption{Unlearn seed 0}
        \end{subfigure}
        \begin{subfigure}[b]{0.9\columnwidth}
            \centering
            \includegraphics[width=\linewidth]{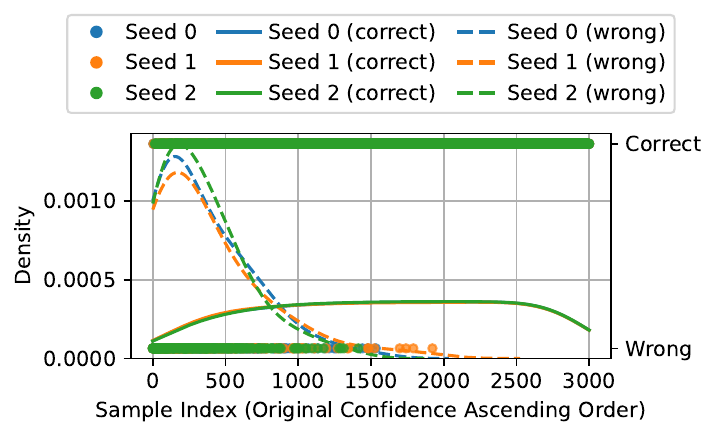}
            \caption{Unlearn seed 1}
        \end{subfigure}
        \begin{subfigure}[b]{0.9\columnwidth}
            \centering
            \includegraphics[width=\linewidth]{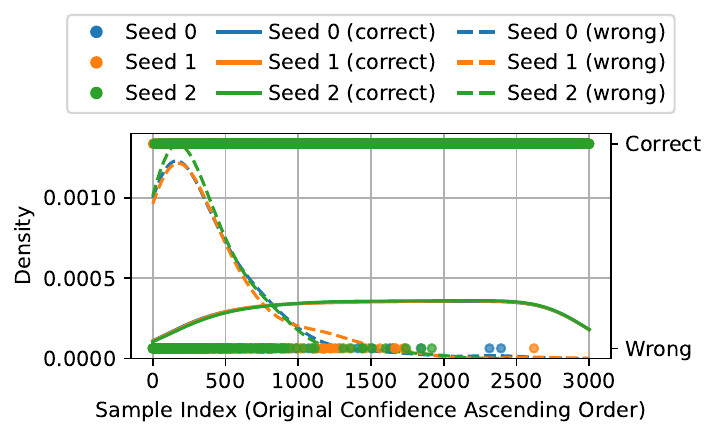}
            \caption{Unlearn seed 2}
        \end{subfigure}
        \caption{CIFAR10-ResNet18 correctness distributions}
        \label{fig:cifar10resnet18correct}
    \end{figure}

\begin{figure}[htbp]
        \centering
        \begin{subfigure}[b]{0.9\columnwidth}
            \centering
            \includegraphics[width=\linewidth]{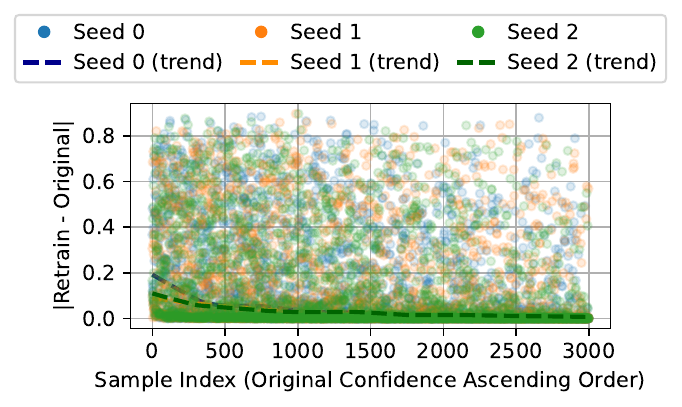}
            \caption{Unlearn seed 0}
        \end{subfigure}
        \begin{subfigure}[b]{0.9\columnwidth}
            \centering
            \includegraphics[width=\linewidth]{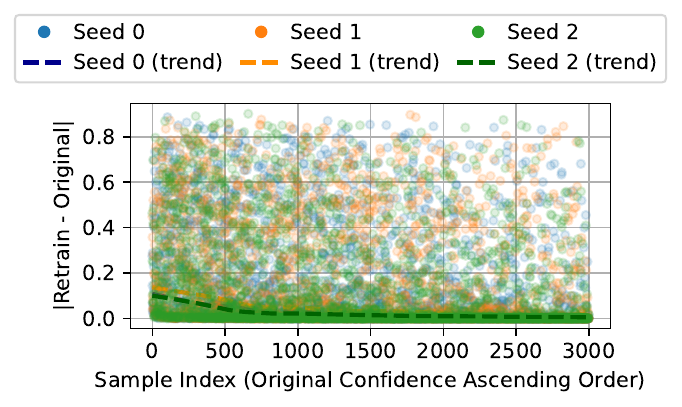}
            \caption{Unlearn seed 1}
        \end{subfigure}
        \begin{subfigure}[b]{0.9\columnwidth}
            \centering
            \includegraphics[width=\linewidth]{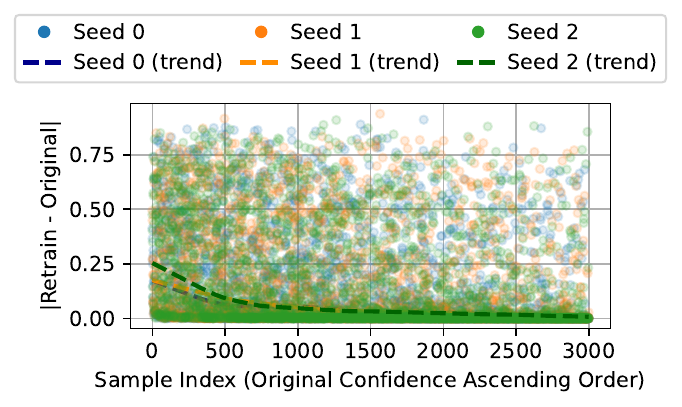}
            \caption{Unlearn seed 2}
        \end{subfigure}
        \caption{CIFAR100-ResNet50 Confidence difference of forget set}
        \label{fig:cifar100resnet50cd}
    \end{figure}

\begin{figure}[htbp]
        \centering
        \begin{subfigure}[b]{0.9\columnwidth}
            \centering
            \includegraphics[width=\linewidth]{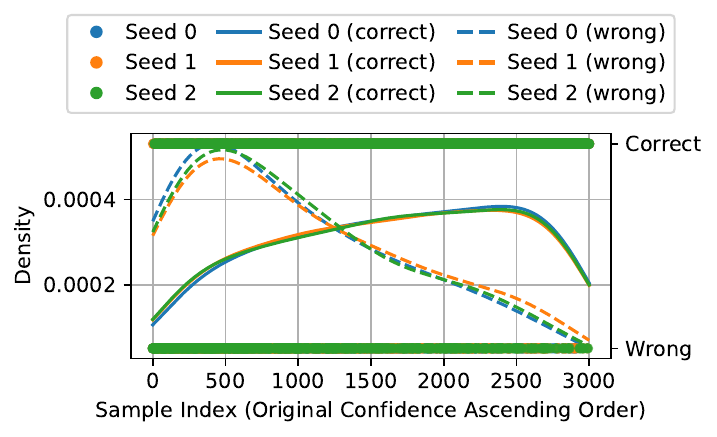}
            \caption{Unlearn seed 0}
        \end{subfigure}
        \begin{subfigure}[b]{0.9\columnwidth}
            \centering
            \includegraphics[width=\linewidth]{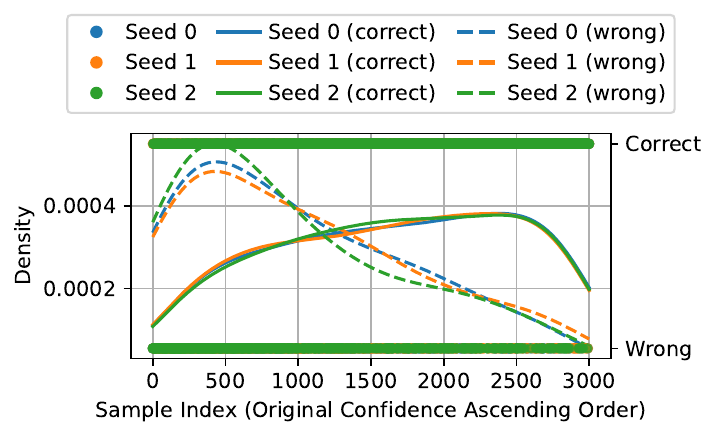}
            \caption{Unlearn seed 1}
        \end{subfigure}
        \begin{subfigure}[b]{0.9\columnwidth}
            \centering
            \includegraphics[width=\linewidth]{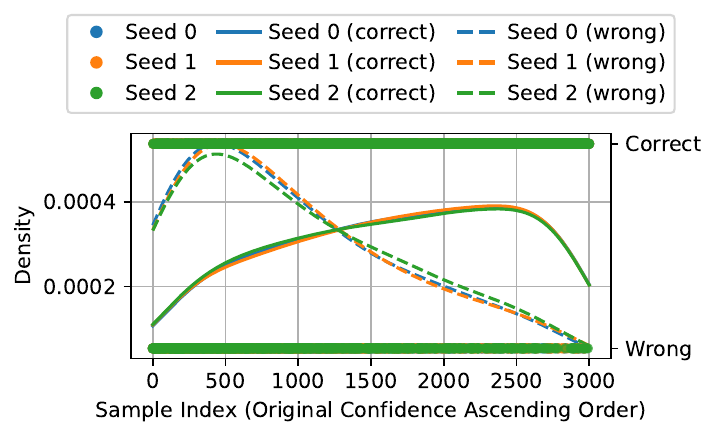}
            \caption{Unlearn seed 2}
        \end{subfigure}
        \caption{CIFAR100-ResNet50 correctness distributions}
        \label{fig:cifar100resnet50correct}
    \end{figure}

\begin{figure}[htbp]
        \centering
        \begin{subfigure}[b]{0.9\columnwidth}
            \centering
            \includegraphics[width=\linewidth]{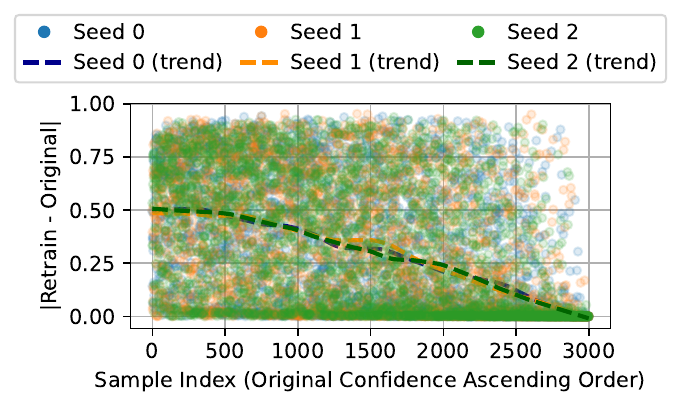}
            \caption{Unlearn seed 0}
        \end{subfigure}
        \begin{subfigure}[b]{0.9\columnwidth}
            \centering
            \includegraphics[width=\linewidth]{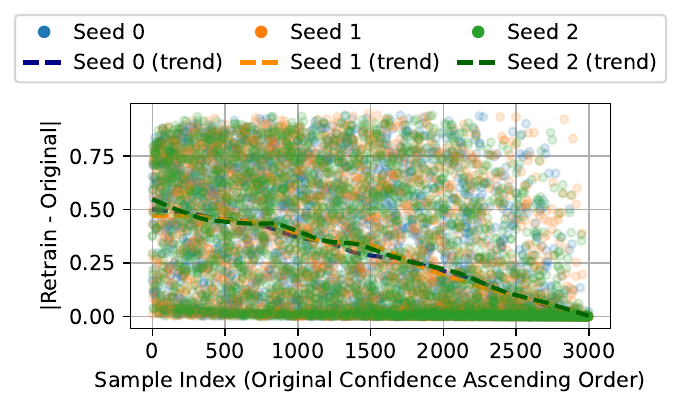}
            \caption{Unlearn seed 1}
        \end{subfigure}
        \begin{subfigure}[b]{0.9\columnwidth}
            \centering
            \includegraphics[width=\linewidth]{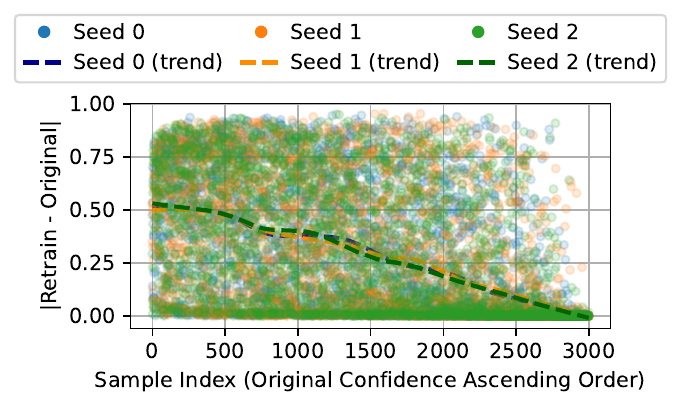}
            \caption{Unlearn seed 2}
        \end{subfigure}
        \caption{TinyImageNet-ResNet18 Confidence difference of forget set}
        \label{fig:TinyImagenetresnet18cd}
    \end{figure}

\begin{figure}[htbp]
        \centering
        \begin{subfigure}[b]{0.9\columnwidth}
            \centering
            \includegraphics[width=\linewidth]{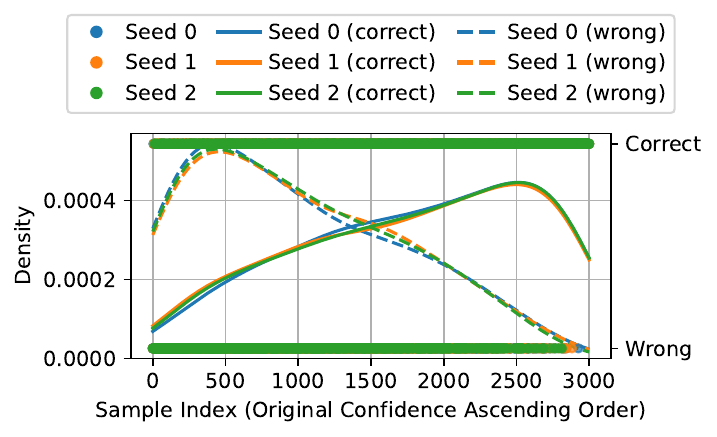}
            \caption{Unlearn seed 0}
        \end{subfigure}
        \begin{subfigure}[b]{0.9\columnwidth}
            \centering
            \includegraphics[width=\linewidth]{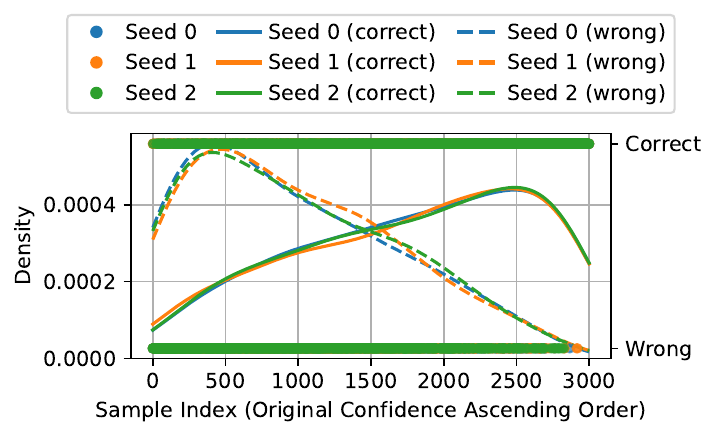}
            \caption{Unlearn seed 1}
        \end{subfigure}
        \begin{subfigure}[b]{0.9\columnwidth}
            \centering
            \includegraphics[width=\linewidth]{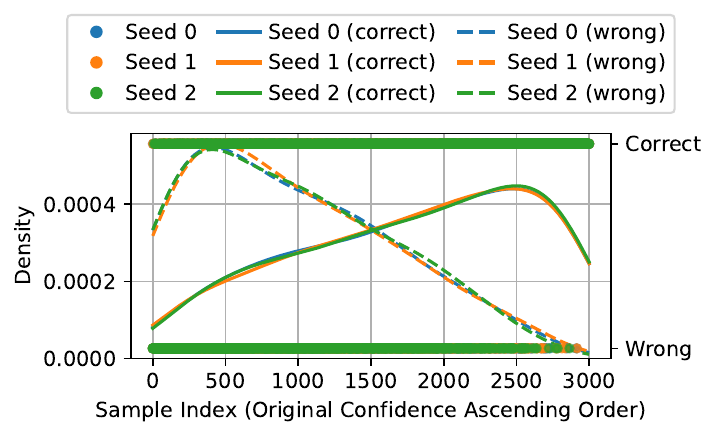}
            \caption{Unlearn seed 2}
        \end{subfigure}
        \caption{TinyImageNet-ResNet18 correctness distributions}
        \label{fig:TinyImagenetresnet18correct}
    \end{figure}

\begin{figure}[htbp]
        \centering
        \begin{subfigure}[b]{0.9\columnwidth}
            \centering
            \includegraphics[width=\linewidth]{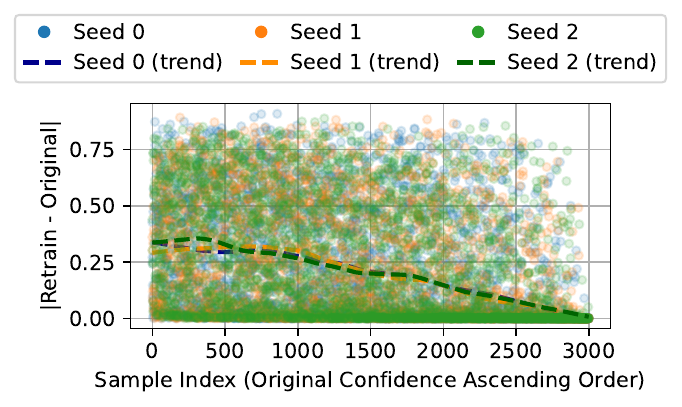}
            \caption{Unlearn seed 0}
        \end{subfigure}
        \begin{subfigure}[b]{0.9\columnwidth}
            \centering
            \includegraphics[width=\linewidth]{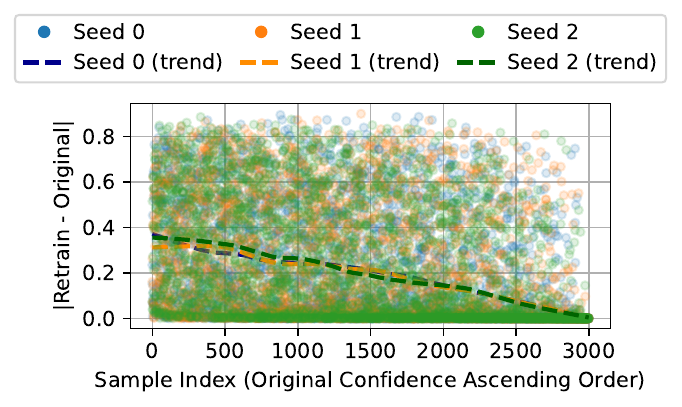}
            \caption{Unlearn seed 1}
        \end{subfigure}
        \begin{subfigure}[b]{0.9\columnwidth}
            \centering
            \includegraphics[width=\linewidth]{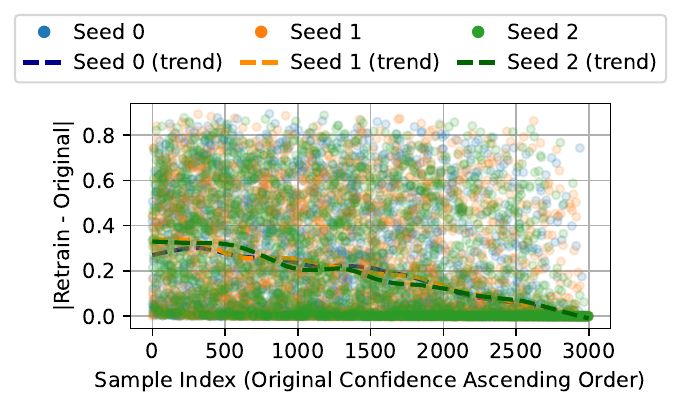}
            \caption{Unlearn seed 2}
        \end{subfigure}
        \caption{TinyImageNet-VGG16-bn Confidence difference of forget set}
        \label{fig:TinyImagenetvgg16_bncd}
    \end{figure}

\begin{figure}[htbp]
        \centering
        \begin{subfigure}[b]{0.9\columnwidth}
            \centering
            \includegraphics[width=\linewidth]{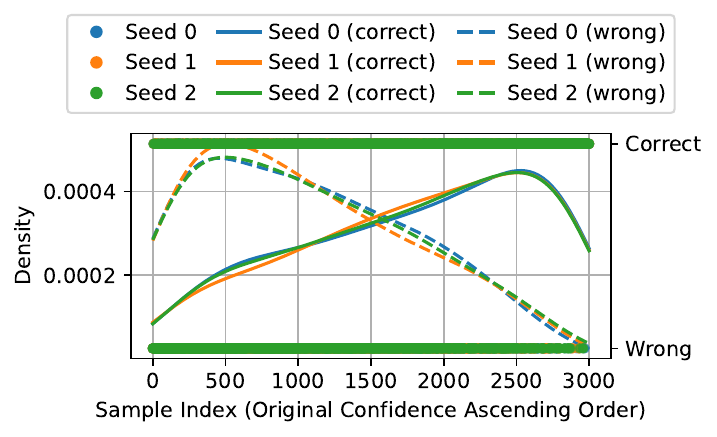}
            \caption{Unlearn seed 0}
        \end{subfigure}
        \begin{subfigure}[b]{0.9\columnwidth}
            \centering
            \includegraphics[width=\linewidth]{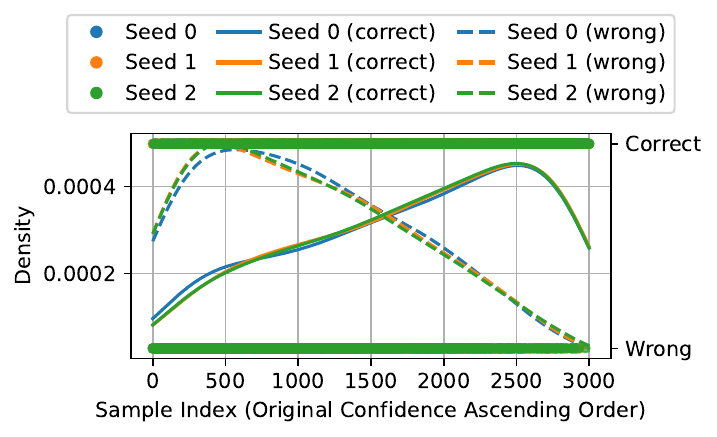}
            \caption{Unlearn seed 1}
        \end{subfigure}
        \begin{subfigure}[b]{0.9\columnwidth}
            \centering
            \includegraphics[width=\linewidth]{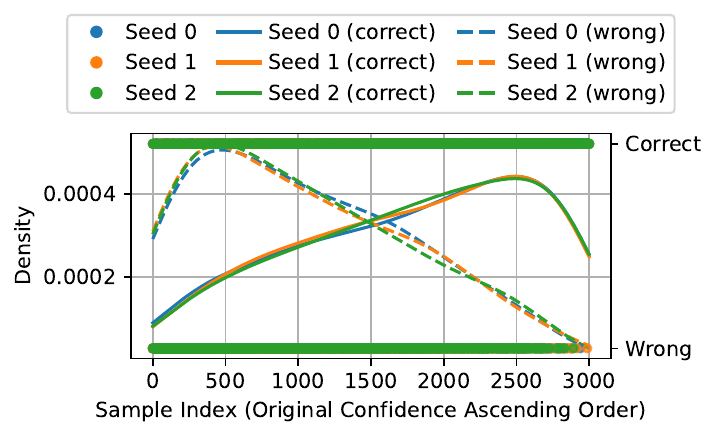}
            \caption{Unlearn seed 2}
        \end{subfigure}
        \caption{TinyImageNet-VGG16-bn correctness distributions}
        \label{fig:TinyImagenetvgg16_bncorrect}
    \end{figure}

\begin{figure*}[htbp]
        \centering
        \begin{subfigure}[b]{0.485\textwidth}
            \centering
            \includegraphics[height=0.38\textheight]{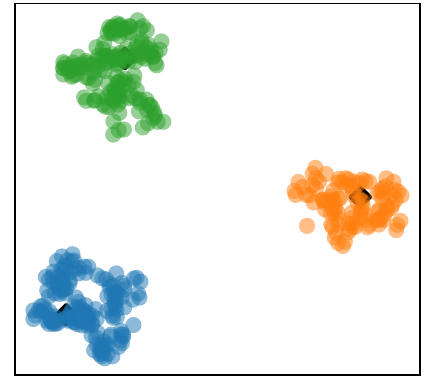}
            \caption{Original model feature with high confidence samples}
            \label{fig:conf-orig-high-cifar10-resnet18}
        \end{subfigure}
        \begin{subfigure}[b]{0.485\textwidth}
            \centering
            \includegraphics[height=0.38\textheight]{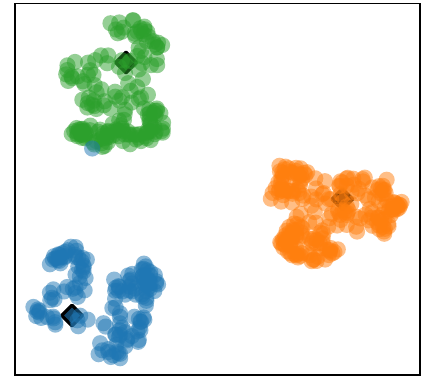}
            \caption{Original model feature with low confidence samples}
            \label{fig:conf-orig-low-cifar10-resnet18}
        \end{subfigure}
        \begin{subfigure}[b]{0.485\textwidth}
            \centering
            \includegraphics[height=0.38\textheight]{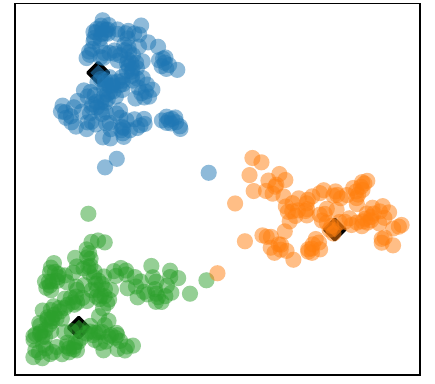}
            \caption{Retrained model feature with high confidence samples}
            \label{fig:conf-retr-high-cifar10-resnet18}
        \end{subfigure}
        \begin{subfigure}[b]{0.485\textwidth}
            \centering
            \includegraphics[height=0.38\textheight]{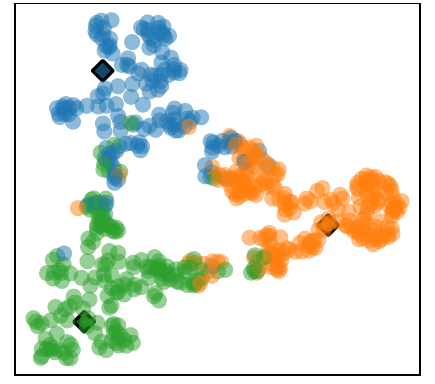}
            \caption{Retrained model feature with low confidence samples}
            \label{fig:conf-retr-low-cifar10-resnet18}
        \end{subfigure}
        \caption{T-SNE analysis of feature space for CIFAR10 and ResNet18}
        \label{fig:confidence-analysis-cifar10-resnet18}
    \end{figure*}

\begin{figure*}[htbp]
        \centering
        \begin{subfigure}[b]{0.485\textwidth}
            \centering
            \includegraphics[height=0.38\textheight]{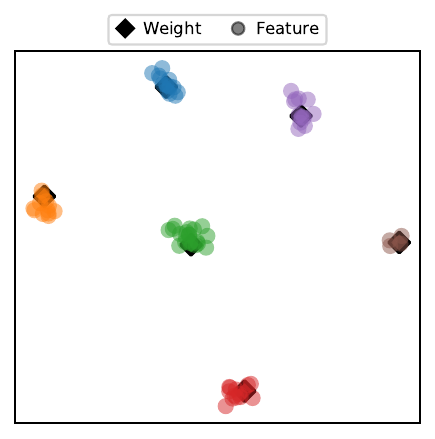}
            \caption{Original model feature with high confidence samples}
            \label{fig:conf-orig-high-cifar100-resnet50}
        \end{subfigure}
        \begin{subfigure}[b]{0.485\textwidth}
            \centering
            \includegraphics[height=0.38\textheight]{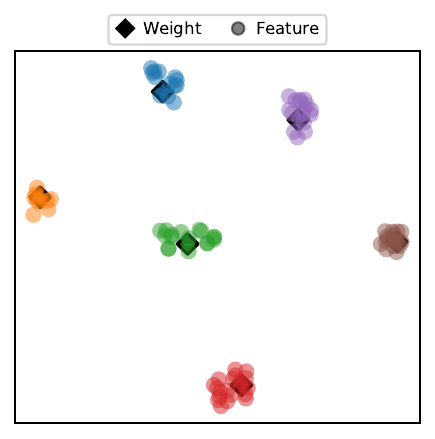}
            \caption{Original model feature with low confidence samples}
            \label{fig:conf-orig-low-cifar100-resnet50}
        \end{subfigure}
        \begin{subfigure}[b]{0.485\textwidth}
            \centering
            \includegraphics[height=0.38\textheight]{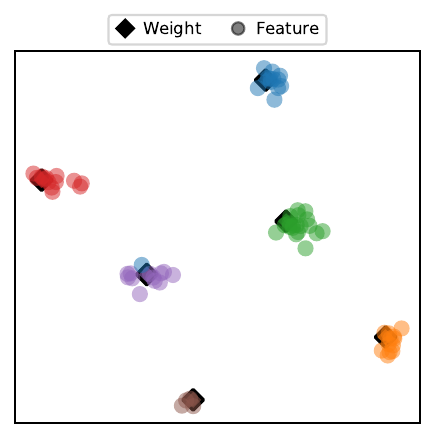}
            \caption{Retrained model feature with high confidence samples}
            \label{fig:conf-retr-high-cifar100-resnet50}
        \end{subfigure}
        \begin{subfigure}[b]{0.485\textwidth}
            \centering
            \includegraphics[height=0.38\textheight]{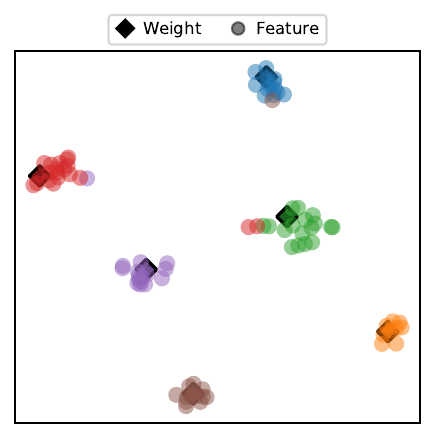}
            \caption{Retrained model feature with low confidence samples}
            \label{fig:conf-retr-low-cifar100-resnet50}
        \end{subfigure}
        \caption{T-SNE analysis of feature space for CIFAR100 and ResNet50}
        \label{fig:confidence-analysis-cifar100-resnet50}
    \end{figure*}

\begin{figure*}[htbp]
        \centering
        \begin{subfigure}[b]{0.485\textwidth}
            \centering
            \includegraphics[height=0.38\textheight]{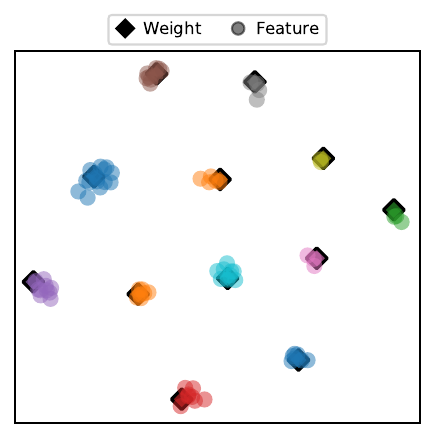}
            \caption{Original model feature with high confidence samples}
            \label{fig:conf-orig-high-TinyImagenet-resnet18}
        \end{subfigure}
        \begin{subfigure}[b]{0.485\textwidth}
            \centering
            \includegraphics[height=0.38\textheight]{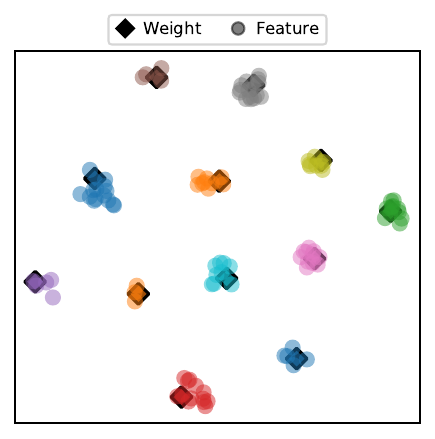}
            \caption{Original model feature with low confidence samples}
            \label{fig:conf-orig-low-TinyImagenet-resnet18}
        \end{subfigure}
        \begin{subfigure}[b]{0.485\textwidth}
            \centering
            \includegraphics[height=0.38\textheight]{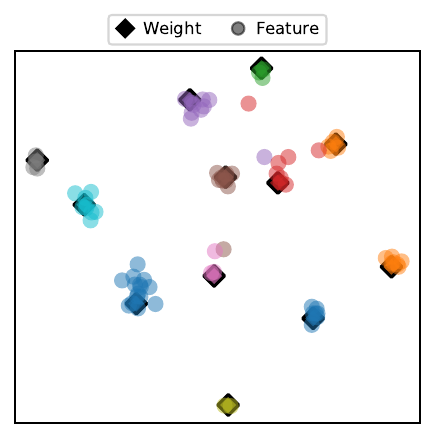}
            \caption{Retrained model feature with high confidence samples}
            \label{fig:conf-retr-high-TinyImagenet-resnet18}
        \end{subfigure}
        \begin{subfigure}[b]{0.485\textwidth}
            \centering
            \includegraphics[height=0.38\textheight]{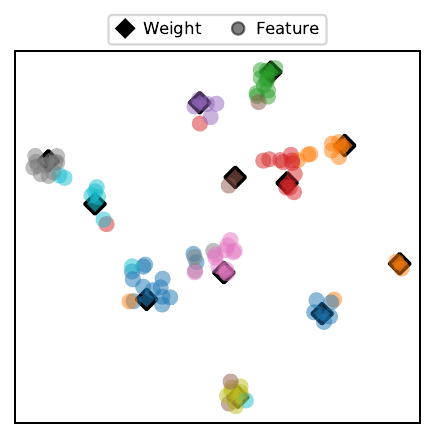}
            \caption{Retrained model feature with low confidence samples}
            \label{fig:conf-retr-low-TinyImagenet-resnet18}
        \end{subfigure}
        \caption{T-SNE analysis of feature space for TinyImageNet and ResNet18}
        \label{fig:confidence-analysis-TinyImagenet-resnet18}
    \end{figure*}

\begin{figure*}[htbp]
        \centering
        \begin{subfigure}[b]{0.485\textwidth}
            \centering
            \includegraphics[height=0.38\textheight]{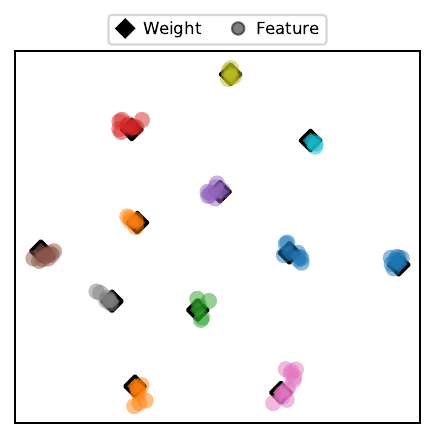}
            \caption{Original model feature with high confidence samples}
            \label{fig:conf-orig-high-TinyImagenet-vgg16_bn}
        \end{subfigure}
        \begin{subfigure}[b]{0.485\textwidth}
            \centering
            \includegraphics[height=0.38\textheight]{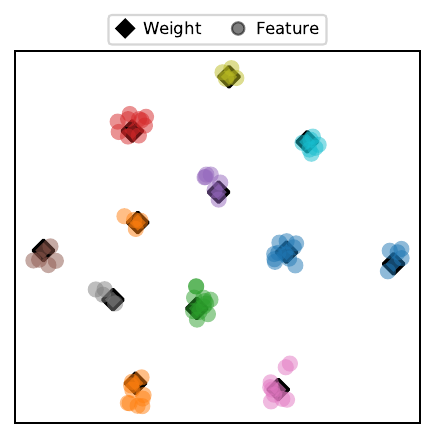}
            \caption{Original model feature with low confidence samples}
            \label{fig:conf-orig-low-TinyImagenet-vgg16_bn}
        \end{subfigure}
        \begin{subfigure}[b]{0.485\textwidth}
            \centering
            \includegraphics[height=0.38\textheight]{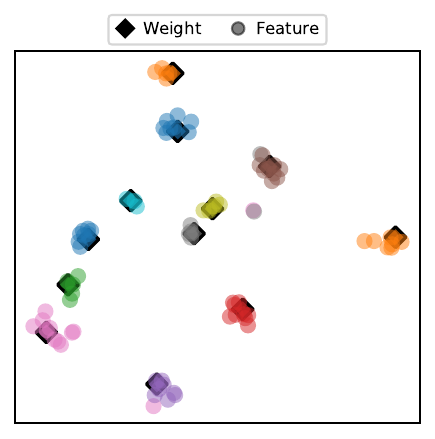}
            \caption{Retrained model feature with high confidence samples}
            \label{fig:conf-retr-high-TinyImagenet-vgg16_bn}
        \end{subfigure}
        \begin{subfigure}[b]{0.485\textwidth}
            \centering
            \includegraphics[height=0.38\textheight]{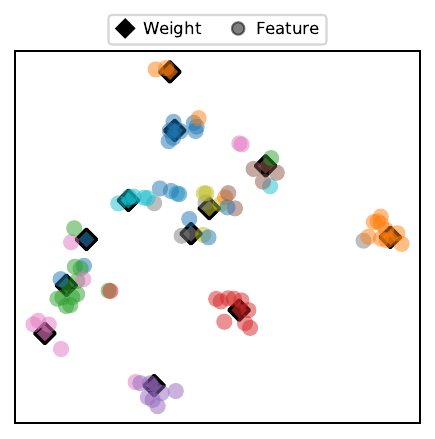}
            \caption{Retrained model feature with low confidence samples}
            \label{fig:conf-retr-low-TinyImagenet-vgg16_bn}
        \end{subfigure}
        \caption{T-SNE analysis of feature space for TinyImageNet and VGG16-bn}
        \label{fig:confidence-analysis-TinyImagenet-vgg16_bn}
    \end{figure*}

In this section, we present additional results of confidence and correctness change across all experimental setups and seeds, focusing on samples in $\mathcal{D}_{f}$ that were considered seen data in $\mathcal{M}_{o}$ but became unseen in $\mathcal{M}_{r}$.
Following the approach in \textit{Observation 2: Discrepancy Analysis on $\mathcal{D}_f$} from the main text, we sort the samples in ascending order of their confidence scores from $\mathcal{M}{orig}$.
The results show that samples with low confidence in $\mathcal{M}_{o}$ are more likely to undergo significant changes in confidence and are more frequently misclassified after unlearning.
For the changes in confidence, refer to \Cref{fig:cifar10resnet18correct}, \Cref{fig:cifar100resnet50correct}, \Cref{fig:TinyImagenetresnet18correct}, and \Cref{fig:TinyImagenetvgg16_bncorrect}.
For the changes in prediction correctness, refer to \Cref{fig:cifar10resnet18cd}, \Cref{fig:cifar100resnet50cd}, \Cref{fig:TinyImagenetresnet18cd}, and \Cref{fig:TinyImagenetvgg16_bncd}.
For T-SNE visualizations for all experimental setups, refer to \Cref{fig:confidence-analysis-cifar10-resnet18}, \Cref{fig:confidence-analysis-cifar100-resnet50}, \Cref{fig:confidence-analysis-TinyImagenet-resnet18}, and \Cref{fig:confidence-analysis-TinyImagenet-vgg16_bn}.

\section{MIA Results}
This section presents results for membership inference attacks (MIA) based not only on confidence in the main text, but also on output probability and entropy, as shown in \Cref{tab:mia_prob} and \Cref{tab:mia_entr}, respectively.
Consistent with the claims in the main text, our method significantly improves the MIA gap of SU.
Furthermore, it achieves performance comparable to or better than that of all other baselines evaluated.

\section{Comparison with Baseline}
\subsection{Quantitative Analysis}
\begin{figure*}[htbp]
        \centering
        \begin{subfigure}[b]{0.32\textwidth}
            \centering
            \includegraphics[width=\linewidth]{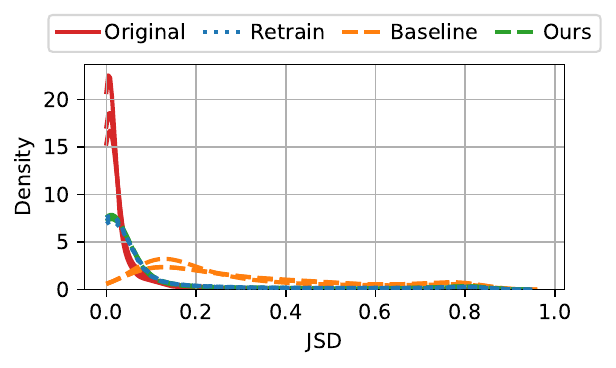}
            \caption{Forget Set}
        \end{subfigure}
        \begin{subfigure}[b]{0.32\textwidth}
            \centering
            \includegraphics[width=\linewidth]{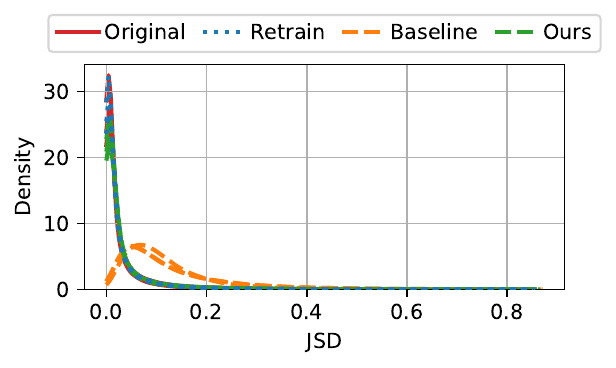}
            \caption{Retain Set}
        \end{subfigure}
        \begin{subfigure}[b]{0.32\textwidth}
            \centering
            \includegraphics[width=\linewidth]{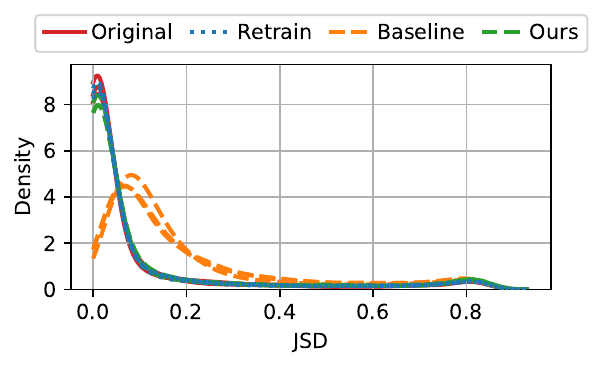}
            \caption{Test Set}
        \end{subfigure}
        \caption{Softmax space visualization for ResNet18 on CIFAR10 (RL)}
        \label{fig:softmax-randomlabel-cifar10-resnet18-seed0}
    \end{figure*}

\begin{figure*}[htbp]
        \centering
        \begin{subfigure}[b]{0.32\textwidth}
            \centering
            \includegraphics[width=\linewidth]{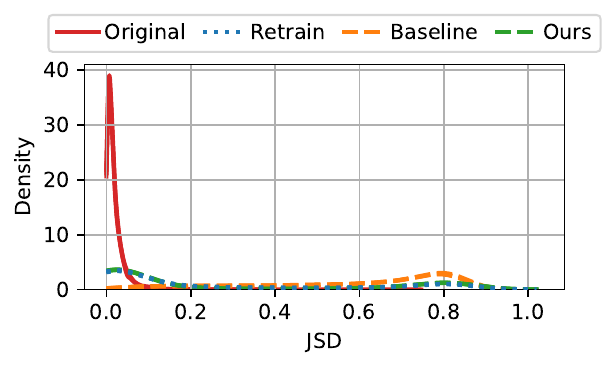}
            \caption{Forget Set}
        \end{subfigure}
        \begin{subfigure}[b]{0.32\textwidth}
            \centering
            \includegraphics[width=\linewidth]{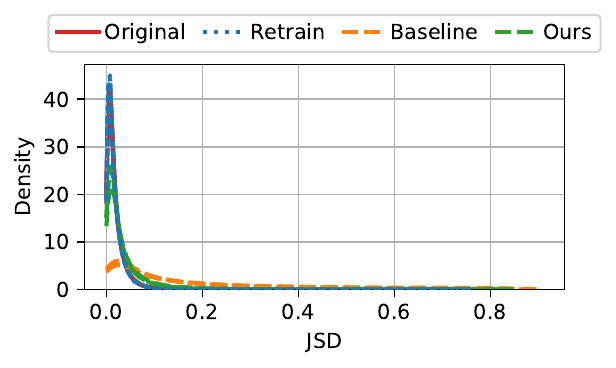}
            \caption{Retain Set}
        \end{subfigure}
        \begin{subfigure}[b]{0.32\textwidth}
            \centering
            \includegraphics[width=\linewidth]{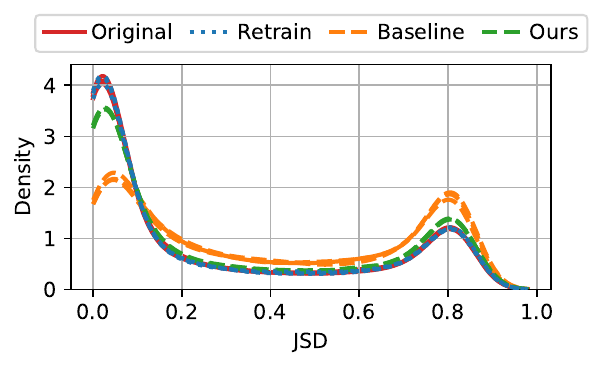}
            \caption{Test Set}
        \end{subfigure}
        \caption{Softmax space visualization for ResNet50 on CIFAR100 (RL)}
        \label{fig:softmax-randomlabel-cifar100-resnet50-seed0}
    \end{figure*}

\begin{figure*}[htbp]
        \centering
        \begin{subfigure}[b]{0.32\textwidth}
            \centering
            \includegraphics[width=\linewidth]{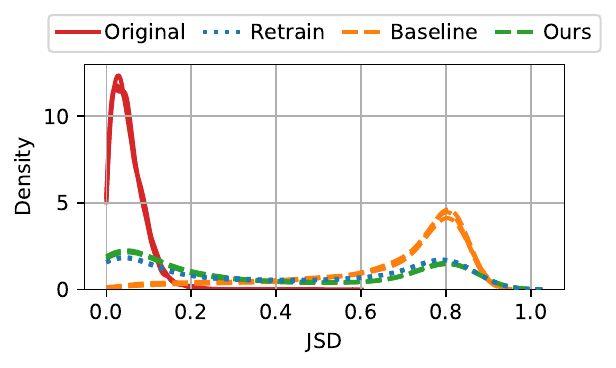}
            \caption{Forget Set}
        \end{subfigure}
        \begin{subfigure}[b]{0.32\textwidth}
            \centering
            \includegraphics[width=\linewidth]{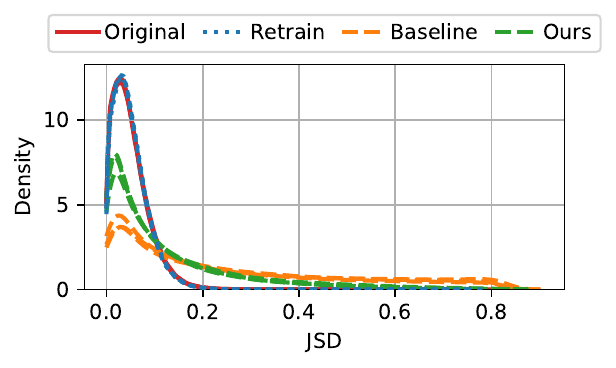}
            \caption{Retain Set}
        \end{subfigure}
        \begin{subfigure}[b]{0.32\textwidth}
            \centering
            \includegraphics[width=\linewidth]{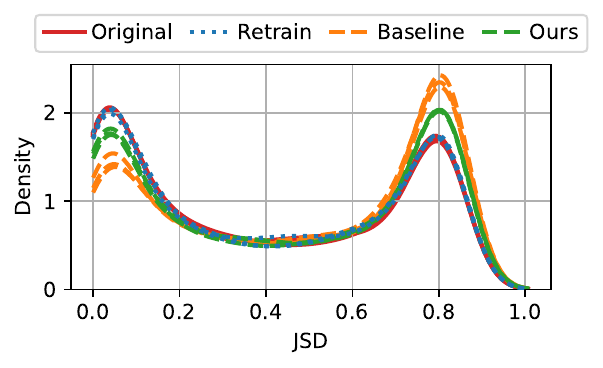}
            \caption{Test Set}
        \end{subfigure}
        \caption{Softmax space visualization for ResNet18 on TinyImageNet (RL)}
        \label{fig:softmax-randomlabel-TinyImagenet-resnet18-seed0}
    \end{figure*}

\begin{figure*}[htbp]
        \centering
        \begin{subfigure}[b]{0.32\textwidth}
            \centering
            \includegraphics[width=\linewidth]{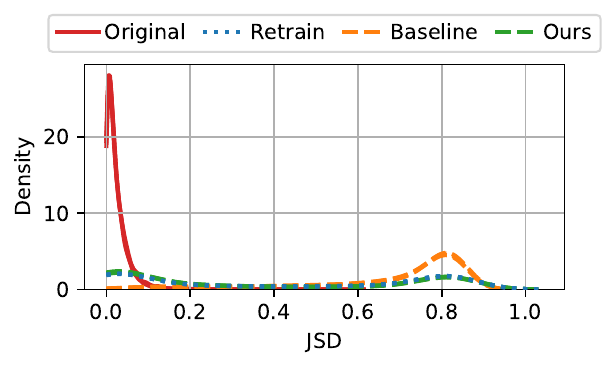}
            \caption{Forget Set}
        \end{subfigure}
        \begin{subfigure}[b]{0.32\textwidth}
            \centering
            \includegraphics[width=\linewidth]{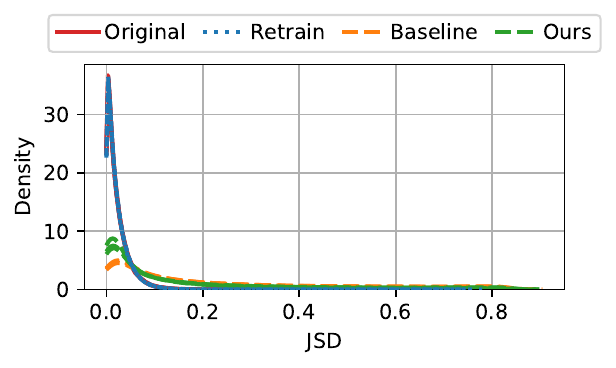}
            \caption{Retain Set}
        \end{subfigure}
        \begin{subfigure}[b]{0.32\textwidth}
            \centering
            \includegraphics[width=\linewidth]{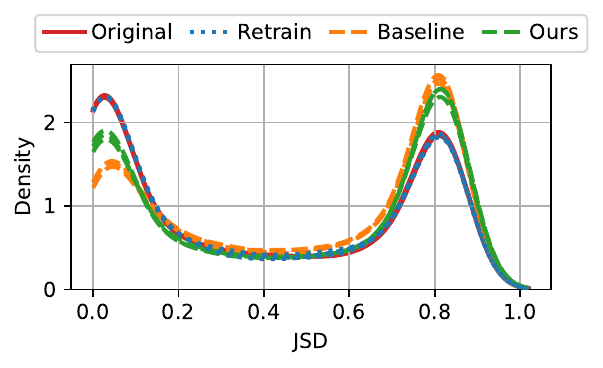}
            \caption{Test Set}
        \end{subfigure}
        \caption{Softmax space visualization for VGG16-bn on TinyImageNet (RL)}
        \label{fig:softmax-randomlabel-TinyImagenet-vgg16_bn-seed0}
    \end{figure*}

\begin{figure*}[htbp]
        \centering
        \begin{subfigure}[b]{0.32\textwidth}
            \centering
            \includegraphics[width=\linewidth]{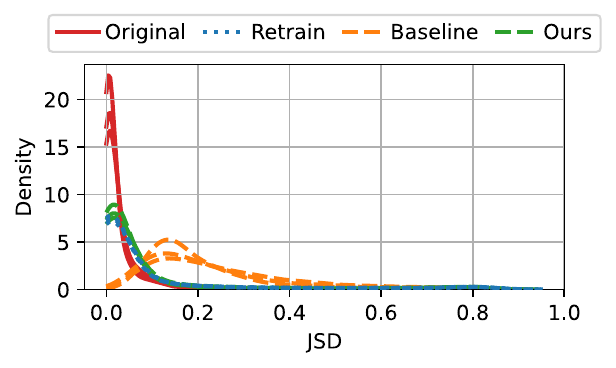}
            \caption{Forget Set}
        \end{subfigure}
        \begin{subfigure}[b]{0.32\textwidth}
            \centering
            \includegraphics[width=\linewidth]{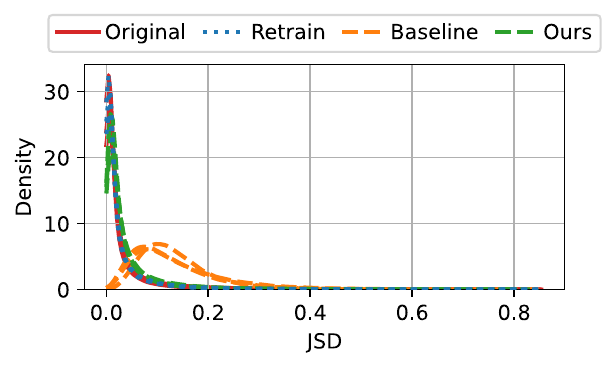}
            \caption{Retain Set}
        \end{subfigure}
        \begin{subfigure}[b]{0.32\textwidth}
            \centering
            \includegraphics[width=\linewidth]{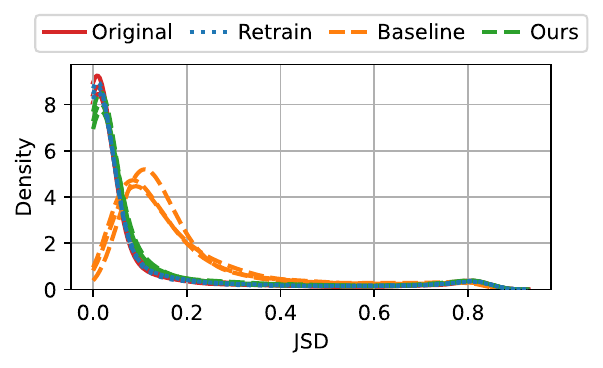}
            \caption{Test Set}
        \end{subfigure}
        \caption{Softmax space visualization for ResNet18 on CIFAR10 (SU)}
        \label{fig:softmax-randomlabel_salun-cifar10-resnet18-seed0}
    \end{figure*}

\begin{figure*}[htbp]
        \centering
        \begin{subfigure}[b]{0.32\textwidth}
            \centering
            \includegraphics[width=\linewidth]{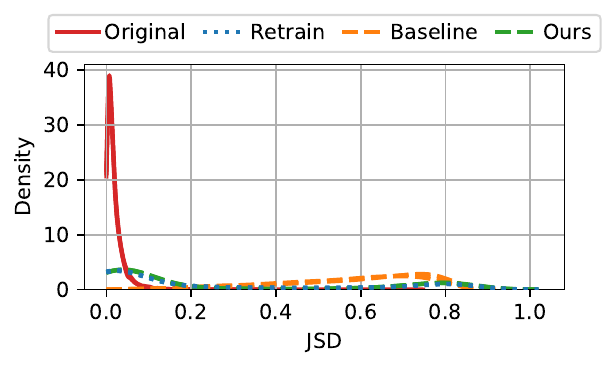}
            \caption{Forget Set}
        \end{subfigure}
        \begin{subfigure}[b]{0.32\textwidth}
            \centering
            \includegraphics[width=\linewidth]{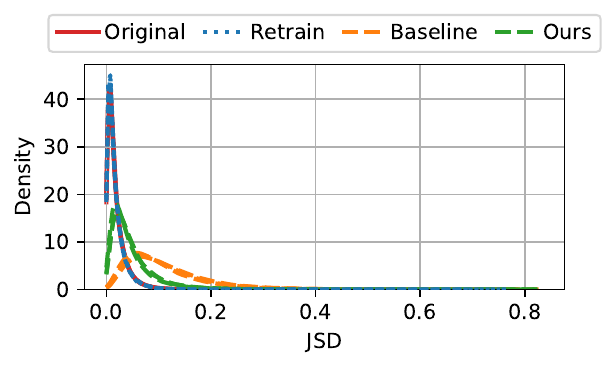}
            \caption{Retain Set}
        \end{subfigure}
        \begin{subfigure}[b]{0.32\textwidth}
            \centering
            \includegraphics[width=\linewidth]{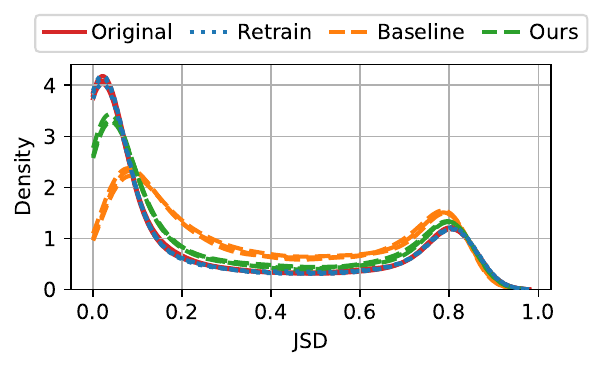}
            \caption{Test Set}
        \end{subfigure}
        \caption{Softmax space visualization for ResNet50 on CIFAR100 (SU)}
        \label{fig:softmax-randomlabel_salun-cifar100-resnet50-seed0}
    \end{figure*}

\begin{figure*}[htbp]
        \centering
        \begin{subfigure}[b]{0.32\textwidth}
            \centering
            \includegraphics[width=\linewidth]{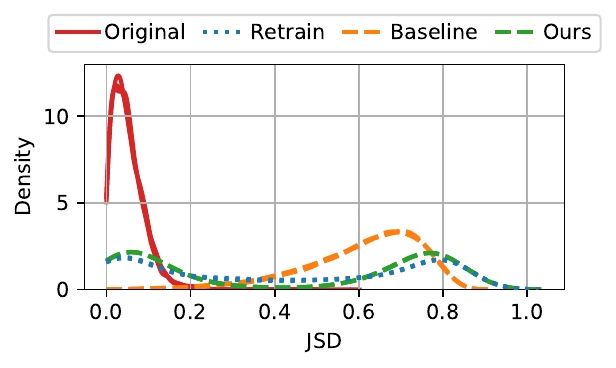}
            \caption{Forget Set}
        \end{subfigure}
        \begin{subfigure}[b]{0.32\textwidth}
            \centering
            \includegraphics[width=\linewidth]{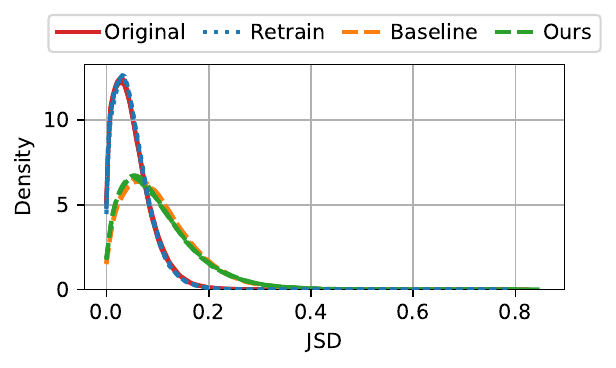}
            \caption{Retain Set}
        \end{subfigure}
        \begin{subfigure}[b]{0.32\textwidth}
            \centering
            \includegraphics[width=\linewidth]{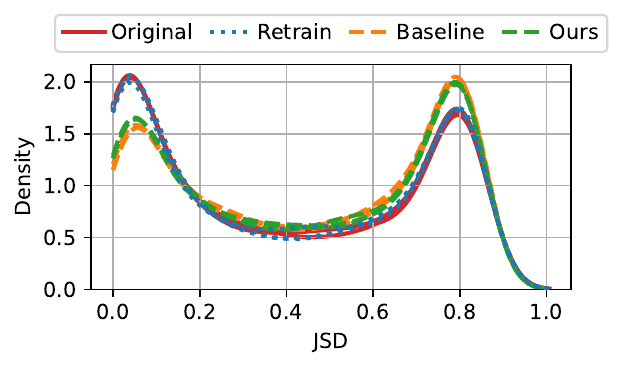}
            \caption{Test Set}
        \end{subfigure}
        \caption{Softmax space visualization for ResNet18 on TinyImageNet (SU)}
        \label{fig:softmax-randomlabel_salun-TinyImagenet-resnet18-seed0}
    \end{figure*}

\begin{figure*}[htbp]
        \centering
        \begin{subfigure}[b]{0.32\textwidth}
            \centering
            \includegraphics[width=\linewidth]{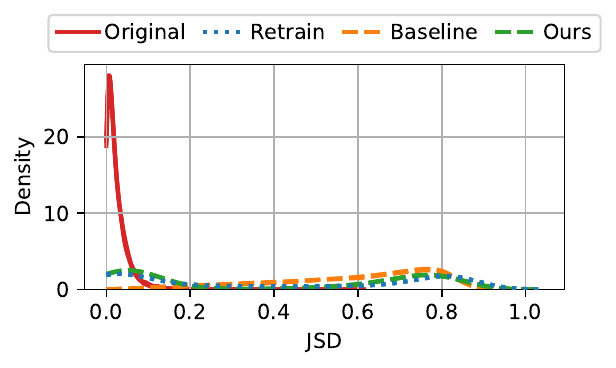}
            \caption{Forget Set}
        \end{subfigure}
        \begin{subfigure}[b]{0.32\textwidth}
            \centering
            \includegraphics[width=\linewidth]{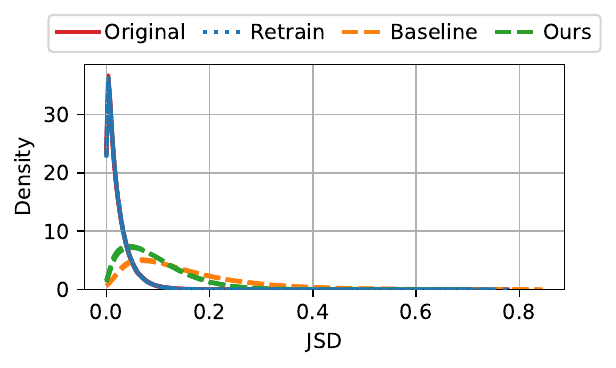}
            \caption{Retain Set}
        \end{subfigure}
        \begin{subfigure}[b]{0.32\textwidth}
            \centering
            \includegraphics[width=\linewidth]{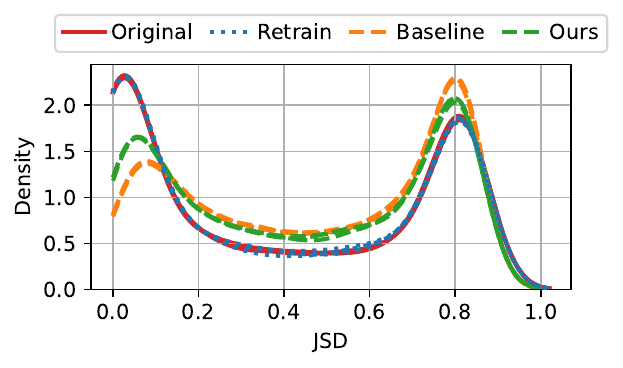}
            \caption{Test Set}
        \end{subfigure}
        \caption{Softmax space visualization for VGG16-bn on TinyImageNet (SU)}
        \label{fig:softmax-randomlabel_salun-TinyImagenet-vgg16_bn-seed0}
    \end{figure*}

This section presents additional comparisons in the Softmax space between our method (RL$+$PeriUn, SU$+$PeriUn) and the baseline (RL, SU).
We observe that PeriUn produces output probabilities more similar to those of the retrained model.
Refer for RL methods \Cref{fig:softmax-randomlabel-cifar10-resnet18-seed0},
\Cref{fig:softmax-randomlabel-cifar100-resnet50-seed0},
\Cref{fig:softmax-randomlabel-TinyImagenet-resnet18-seed0}, and 
\Cref{fig:softmax-randomlabel-TinyImagenet-vgg16_bn-seed0}.
Refer for SU methods \Cref{fig:softmax-randomlabel_salun-cifar10-resnet18-seed0},
\Cref{fig:softmax-randomlabel_salun-cifar100-resnet50-seed0},
\Cref{fig:softmax-randomlabel_salun-TinyImagenet-resnet18-seed0}, and 
\Cref{fig:softmax-randomlabel_salun-TinyImagenet-vgg16_bn-seed0}.

\subsection{Qualitative Comparison}
\begin{figure*}[htbp]
        \centering
      
        \begin{subfigure}{0.23\textwidth}
          \centering
          \includegraphics[width=\linewidth]{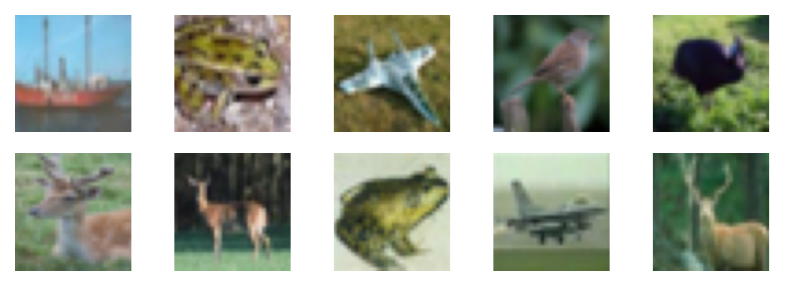}
          \caption{CIFAR10 / ResNet18}
        \end{subfigure}
        \hfill
        \begin{subfigure}{0.23\textwidth}
          \centering
          \includegraphics[width=\linewidth]{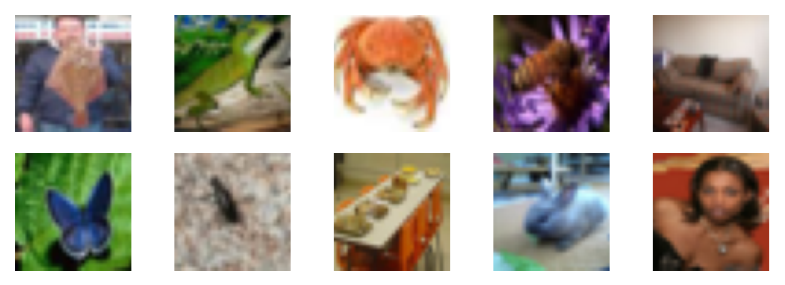}
          \caption{CIFAR100 / ResNet50}
        \end{subfigure}
        \hfill
        \begin{subfigure}{0.23\textwidth}
          \centering
          \includegraphics[width=\linewidth]{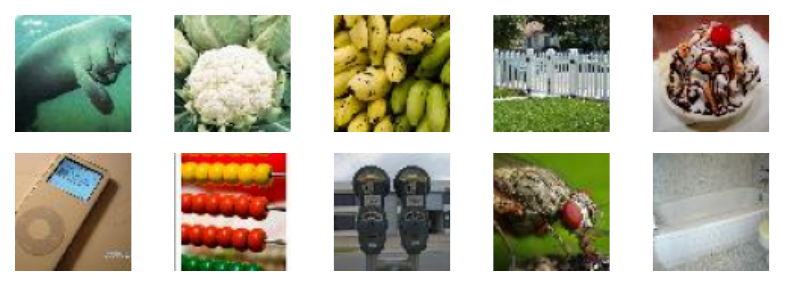}
          \caption{TinyImageNet / ResNet18}
        \end{subfigure}
        \hfill
        \begin{subfigure}{0.23\textwidth}
          \centering
          \includegraphics[width=\linewidth]{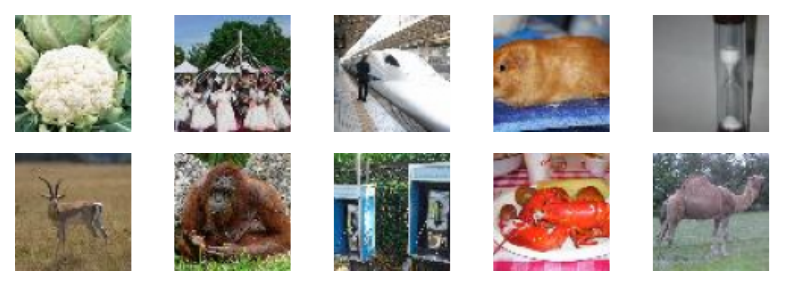}
          \caption{TinyImageNet / VGG16-bn}
        \end{subfigure}
      
        \vspace{1em}
      
        \begin{subfigure}{0.23\textwidth}
          \centering
          \includegraphics[width=\linewidth]{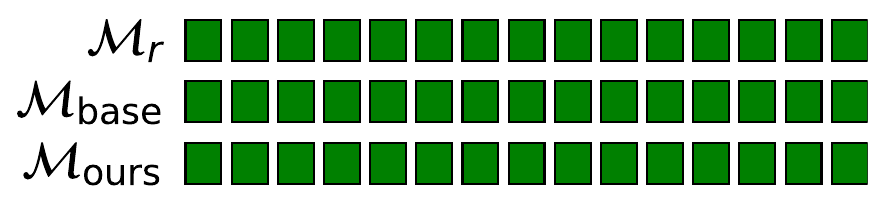}
          \caption{CIFAR10 / ResNet18}
        \end{subfigure}
        \hfill
        \begin{subfigure}{0.23\textwidth}
          \centering
          \includegraphics[width=\linewidth]{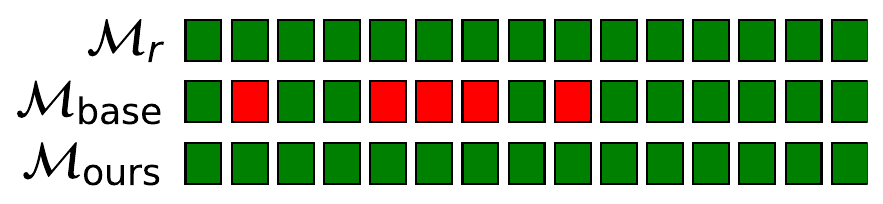}
          \caption{CIFAR100 / ResNet50}
        \end{subfigure}
        \hfill
        \begin{subfigure}{0.23\textwidth}
          \centering
          \includegraphics[width=\linewidth]{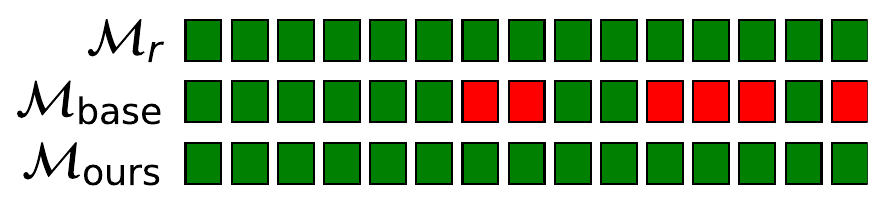}
          \caption{TinyImageNet / ResNet18}
        \end{subfigure}
        \hfill
        \begin{subfigure}{0.23\textwidth}
          \centering
          \includegraphics[width=\linewidth]{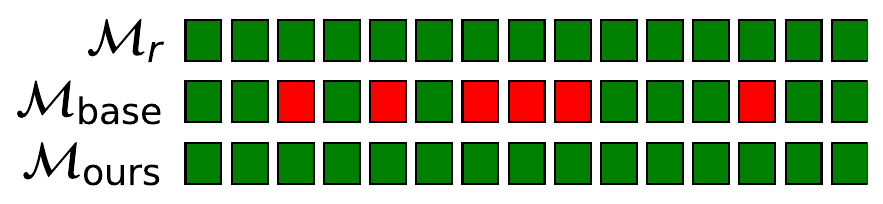}
          \caption{TinyImageNet / VGG16-bn}
        \end{subfigure}
      
        \caption{Qualitative comparison between PeriUn and Baseline (RL). (Above) Top 10 most confident samples selected, mostly visually normal. Images are ordered from left to right, top to bottom, in decreasing order of confidence. (Below) Top 15 highest confident samples are highlighted in red for originally correct and later flipped to incorrect (i.e., deletion) by each model. Retrain and Periun tend to preserve high-confident, while Baseline deletes all requested samples, including high-confidence ones that do not need to be removed, which suggests catastrophic forgetting.
      }
        \label{fig:random_label_4x2}
      \end{figure*}

\Cref{fig:random_label_4x2} presents a comparison between the basic Random Label method and the Random Label method combined with PeriUn.
In addition, we quantitatively evaluated, in the SU setting, the disagreement in correct/incorrect predictions between $(\mathcal{M}_r, \mathcal{M}_{\text{base}})$ and $(\mathcal{M}_r, \mathcal{M}_{\text{ours}})$ for the samples that were not removed during the training of $\mathcal{M}_{\text{ours}}$. The results are summarized in \cref{tab:addi_quan_diff}. Our method aligns more closely with the excluded samples than the baseline.

\begin{listing*}[htbp]%
    \caption{PeriUn: Low-confidence based forget set selection}%
    \label{lst:lowconf-forget}%
    \begin{lstlisting}[language=Python]
    # Calculate sample confidence from the forget set
    original_forget_set = subset(training_data, forget_set_idx)
    logits = model(original_forget_set)
    confidences = softmax(logits, dim=1).max(dim=1).values
    
    # Identify samples with the lowest confidence
    num_to_forget = int(alpha * len(confidences))
    low_confidence_indices = argsort(confidences)[:num_to_forget]
    
    # Construct the final forget set
    selected_forget_set = subset(original_forget_set, low_confidence_indices)
    \end{lstlisting}
    \end{listing*}

    \begin{table*}[htbp]
    \centering
    {\fontsize{9}{11}\selectfont
    \begin{tabular}{lllllll}
    \hline
    Setup & total epochs & Optimizer & learning rate (lr) & lr scheduler                   & moentum & weight decay \\ \hline
    1     & 30           & SGD       & 0.1                & Cosine Annealing               & 0.9     & 5E-04        \\
    2     & 150          & SGD       & 0.1                & Multi Step({[}60, 120{]}, 0.2) & 0.9     & 5E-04        \\
    3     & 80           & SGD       & 0.1                & Cosine Annealing               & 0.9     & 5E-04        \\
    4     & 100          & SGD       & 0.1                & Cosine Annealing               & 0.9     & 5E-04        \\ \hline
    \end{tabular}
    }
    \caption{Optimization hyperparameter for $\mathcal{M}_o$ and $\mathcal{M}_r$}
    \label{tab:hyper1}
    \end{table*}
    
    \begin{table*}[htbp]
    \centering
    {\fontsize{9}{11}\selectfont
    \begin{tabular}{llllll}
    \hline
    Setup & batch size & normalization mean           & normalization std            & Train transform 1          & Train transform 2    \\ \hline
    1     & 256        & {[}0.4914, 0.4822, 0.4465{]} & {[}0.2470, 0.2435, 0.2616{]} & RandomCrop(32, padding=4)  & RandomHorizontalFlip \\
    2     & 256        & {[}0.5071, 0.4866, 0.4409{]} & {[}0.2673, 0.2564, 0.2762{]} & RandomCrop(32, padding=4), & RandomHorizontalFlip \\
    3     & 256        & {[}0.485, 0.456, 0.406{]}    & {[}0.229, 0.224, 0.225{]}    & RandomCrop(64, padding=4)  & RandomHorizontalFlip \\
    4     & 256        & {[}0.485, 0.456, 0.406{]}    & {[}0.229, 0.224, 0.225{]}    & RandomCrop(64, padding=4)  & RandomHorizontalFlip \\ \hline
    \end{tabular}
    }
    \caption{Data preprocessing hyperparameter for $\mathcal{M}_o$ and $\mathcal{M}_r$}
    \label{tab:hyper2}
    \end{table*}

    \begin{table*}[htbp]
        \centering
        \begin{tabular}{llll}
        \hline
        Method &  \multicolumn{1}{c}{$\alpha$}   &  \multicolumn{1}{c}{lr}                                & Addi. Param.                    \\ \hline
        FT     &  \{0.1, 0.3, 0.5 0.7\} & logspace(-5, -1, num=10, base=10) &                                   \\
        FT+L1  &  \{0.1, 0.3, 0.5 0.7\} & logspace(-6, -4, num=10, base=10) & logspace(-5, -1, num=10, base=10) \\
        GAGD   &  \{0.1, 0.3, 0.5 0.7\} & linspace(0.85, 0.99, num=10)      & logspace(-5, -1, num=10, base=10) \\
        GA     &  \{0.1, 0.3, 0.5 0.7\} & logspace(-5, -1, num=10, base=10) &                                   \\
        RL     &  \{0.1, 0.3, 0.5 0.7\} & linspace(0.01, 0.1, num=10)       &                                   \\
        SalUn  &  \{0.1, 0.3, 0.5 0.7\} & linspace(0.01, 0.05, num=10)      & \{0.3, 0.4, 0.5, 0.6, 0.7\}       \\ \hline
        \end{tabular}
        \caption{Hyperparameter search range}
        \label{tab:hyperparam_searchrange}
    \end{table*}

    \begin{table*}[htbp]
        \centering
        \begin{tabular}{lccll}
        \hline
        Method & Setup & $\alpha$ & lr                      & Addi. Param.          \\ \hline
        FT     & 1     & 1.0             & 0.012915496650148800    &                         \\
        FT     & 2     & 1.0             & 0.03593813663804630     &                         \\
        FT     & 3     & 1.0             & 0.012915496650148800    &                         \\
        FT     & 4     & 1.0             & 0.012915496650148800    &                         \\ \hline
        FT+L1  & 1     & 1.0             & 0.012915496650148800    & 0.000012915496650148800 \\
        FT+L1  & 2     & 1.0             & 0.000002782559402207130 & 0.0359381366380463      \\
        FT+L1  & 3     & 1.0             & 0.012915496650148800    & 0.000021544346900318800 \\
        FT+L1  & 4     & 1.0             & 0.012915496650148800    & 0.000012915496650148800 \\ \hline
        GA     & 1     & 1.0             & 0.00007742636826811280  &                         \\
        GA     & 2     & 1.0             & 0.0016681005372000600   &                         \\
        GA     & 3     & 1.0             & 0.0016681005372000600   &                         \\
        GA     & 4     & 1.0             & 0.004641588833612780    &                         \\ \hline
        GAGD   & 1     & 1.0             & 0.004641588833612780    & 0.9277777777777780      \\
        GAGD   & 2     & 1.0             & 0.004641588833612780    & 0.8811111111111110      \\
        GAGD   & 3     & 1.0             & 0.004641588833612780    & 0.9588888888888890      \\
        GAGD   & 4     & 1.0             & 0.001668100537200060    & 0.8811111111111110      \\ \hline
        RL     & 1     & 1.0             & 0.01                    &                         \\
        RL     & 2     & 1.0             & 0.02                    &                         \\
        RL     & 3     & 1.0             & 0.02333333333333330     &                         \\
        RL     & 4     & 1.0             & 0.02                    &                         \\ \hline
        SalUn  & 1     & 1.0             & 0.0188888888888888      & 0.4                     \\
        SalUn  & 2     & 1.0             & 0.0233333333333333      & 0.3                     \\
        SalUn  & 3     & 1.0             & 0.01                    & 0.5                     \\
        SalUn  & 4     & 1.0             & 0.01                    & 0.3                    \\ \hline
        \end{tabular}
        \caption{Reproduction hyperparam}
        \label{tab:opt_hyperparam}
        \end{table*}

    \begin{table*}[htbp]
        \centering
        \begin{tabular}{lccll}
        \hline
        method & setup & $\alpha$ & lr                          & Addi. Param. \\ \hline
        SalUn  & 1     & \textbf{0.1}    & \textbf{0.0188888888888888} & \textbf{0.7}   \\
        SalUn  & 1     & 0.3             & 0.01                        & 0.6            \\
        SalUn  & 1     & 0.5             & 0.0233333333333333          & 0.3            \\
        SalUn  & 1     & 0.7             & 0.01                        & 0.6            \\ \hline
        SalUn  & 2     & 0.1             & 0.0277777777777777          & 0.7            \\
        SalUn  & 2     & \textbf{0.3}    & \textbf{0.0188888888888888} & \textbf{0.7}   \\
        SalUn  & 2     & 0.5             & 0.0233333333333333          & 0.4            \\
        SalUn  & 2     & 0.7             & 0.0366666666666666          & 0.3            \\ \hline
        SalUn  & 3     & 0.1             & 0.0277777777777777          & 0.7            \\
        SalUn  & 3     & 0.3             & 0.01                        & 0.7            \\
        SalUn  & 3     & \textbf{0.5}    & \textbf{0.01}               & \textbf{0.6}   \\
        SalUn  & 3     & 0.7             & 0.0144444444444444          & 0.5            \\ \hline
        SalUn  & 4     & 0.1             & 0.0277777777777777          & 0.7            \\
        SalUn  & 4     & 0.3             & 0.01                        & 0.7            \\
        SalUn  & 4     & \textbf{0.5}    & \textbf{0.01}               & \textbf{0.3}   \\
        SalUn  & 4     & 0.7             & 0.01                        & 0.3           \\ \hline
        \end{tabular}
        \caption{SalUn hyperparam abilation}
        \label{tab:optsu_hyperparam}
        \end{table*}

        \begin{table*}[htbp]
        \centering
        \begin{tabular}{lccllc}
        \hline
        Method & Setup & $\alpha$ & lr                                            & ToW                      \\ \hline
        GA     & 1     & 0.1             & 0.00007742636826811280                 & 92.46$\pm$0.67           \\
        GA     & 1     & 0.3             & 0.00007742636826811280                 & 92.34$\pm$0.72           \\
        GA     & 1     & 0.5             & 0.0005994842503189410                  & 92.49$\pm$0.79           \\
        GA     & 1     & 0.7             & 0.0005994842503189410                  & \textbf{92.77$\pm$0.65}  \\ \hline
        GA     & 2     & 0.1             & 0.00001                                & 75.37$\pm$0.54           \\
        GA     & 2     & 0.3             & 0.00002782559402207130                 & 75.57$\pm$0.72  \\
        GA     & 2     & 0.5             & 0.004641588833612780                   & 75.22$\pm$0.97           \\
        GA     & 2     & 0.7             & 0.001668100537200060                   & \textbf{75.73$\pm$0.89}  \\ \hline
        GA     & 3     & 0.1             & 0.00021544346900318800                 & 64.09$\pm$1.13           \\
        GA     & 3     & 0.3             & 0.001668100537200060                   & 64.06$\pm$1.16           \\
        GA     & 3     & 0.5             & 0.001668100537200060                   & \textbf{64.11$\pm$1.10}  \\
        GA     & 3     & 0.7             & 0.001668100537200060                   & 64.10$\pm$1.19           \\ \hline
        GA     & 4     & 0.1             & 0.0016681005372                        & 59.79$\pm$1.35           \\
        GA     & 4     & 0.3             & 0.0046415888336127                     & \textbf{59.91$\pm$1.35}  \\
        GA     & 4     & 0.5             & 0.0046415888336127                     & 59.84$\pm$1.28           \\
        GA     & 4     & 0.7             & 0.0046415888336127                     & 53.03$\pm$19.89 \\ \hline
        \end{tabular}
        \caption{GA+ours}
        \label{tab:gapours}
        \end{table*}

        \begin{table*}[htbp]
        \centering
        \begin{tabular}{lccllc}
        \hline
        method & setup & $\alpha$ & lr                    & Addi. Param.     & ToW                     \\ \hline
        GAGD   & 1     & 0.1             & 0.004641588833612780  & 0.99               & 95.98$\pm$0.53          \\
        GAGD   & 1     & 0.3             & 0.012915496650148800  & 0.99               & 95.89$\pm$0.96          \\
        GAGD   & 1     & 0.5             & 0.001668100537200060  & 0.9277777777777780 & 96.55$\pm$0.59          \\
        GAGD   & 1     & 0.7             & 0.004641588833612780  & 0.9588888888888890 & \textbf{96.64$\pm$0.75} \\ \hline
        GAGD   & 2     & 0.1             & 0.004641588833612780  & 0.9588888888888890 & 81.75$\pm$0.66          \\
        GAGD   & 2     & 0.3             & 0.004641588833612780  & 0.9433333333333330 & \textbf{93.00$\pm$1.55} \\
        GAGD   & 2     & 0.5             & 0.004641588833612780  & 0.9277777777777780 & 91.06$\pm$2.14          \\
        GAGD   & 2     & 0.7             & 0.004641588833612780  & 0.9122222222222220 & 89.07$\pm$1.88          \\ \hline
        GAGD   & 3     & 0.1             & 0.012915496650148800  & 0.99               & 72.20$\pm$2.07          \\
        GAGD   & 3     & 0.3             & 0.004641588833612780  & 0.9744444444444440 & 89.80$\pm$1.56          \\
        GAGD   & 3     & 0.5             & 0.001668100537200060  & 0.9433333333333330 & \textbf{93.07$\pm$1.71} \\
        GAGD   & 3     & 0.7             & 0.0005994842503189410 & 0.8655555555555560 & 90.70$\pm$1.34          \\ \hline
        GAGD   & 4     & 0.1             & 0.0005994842503189    & 0.958888888888889  & 64.72$\pm$1.46          \\
        GAGD   & 4     & 0.3             & 0.0129154966501488    & 0.99               & 75.68$\pm$1.40          \\
        GAGD   & 4     & 0.5             & 0.0016681005372       & 0.927777777777778  & 67.36$\pm$36.65         \\
        GAGD   & 4     & 0.7             & 0.0016681005372       & 0.927777777777778  & \textbf{76.78$\pm$8.83} \\ \hline
        \end{tabular}
        \caption{GAGD+ours}
        \label{tab:gagdpours}
        \end{table*}

        \begin{table*}[htbp]
        \centering
        \begin{tabular}{lcccc}
        \hline
        Method & setup & $\alpha$ & lr                    & ToW                      \\ \hline
        RL     & 1     & 0.1             & 0.01                 & \textbf{97.76$\pm$0.43} \\
        RL     & 1     & 0.3             & 0.01                 & 89.75$\pm$0.95          \\
        RL     & 1     & 0.5             & 0.01                 & 84.49$\pm$2.05          \\
        RL     & 1     & 0.7             & 0.01                 & 82.46$\pm$1.65 \\ \hline
        RL     & 2     & 0.1             & 0.01                 & 83.96$\pm$0.61          \\
        RL     & 2     & 0.3             & 0.01                 & \textbf{95.65$\pm$1.46} \\
        RL     & 2     & 0.5             & 0.01                 & 78.87$\pm$1.45          \\
        RL     & 2     & 0.7             & 0.02                 & 65.29$\pm$1.61 \\ \hline
        RL     & 3     & 0.1             & 0.02                 & 81.52$\pm$0.97          \\
        RL     & 3     & 0.3             & 0.01                 & \textbf{86.90$\pm$1.88} \\
        RL     & 3     & 0.5             & 0.01                 & 80.43$\pm$1.61 \\
        RL     & 3     & 0.7             & 0.01                 & 68.16$\pm$1.15          \\ \hline
        RL     & 4     & 0.1             & 0.02                 & 74.73$\pm$1.94          \\
        RL     & 4     & 0.3             & 0.01                 & \textbf{79.99$\pm$1.91} \\
        RL     & 4     & 0.5             & 0.01                 & 75.14$\pm$1.96          \\
        RL     & 4     & 0.7             & 0.01                 & 63.83$\pm$1.35 \\ \hline
        \end{tabular}
        \caption{RL+ours}
        \label{tab:rlpours}
        \end{table*}

    \begin{table*}
        \centering
        \begin{tabular}{l|cc|cc} \hline
                   & \multicolumn{2}{c}{Setup1}      & \multicolumn{2}{c}{Setup2}     \\
                   & \multicolumn{1}{c}{MIA}      & \multicolumn{1}{c}{Gap}    & \multicolumn{1}{c}{MIA}     & \multicolumn{1}{c}{Gap}     \\
        Retrain    & 13.22$\pm$4.17               & \multicolumn{1}{l}{}           & 60.32$\pm$3.87              & \multicolumn{1}{l}{}            \\
        FT         & 14.94$\pm$7.74               & \textbf{1.72}                  & 41.90$\pm$5.52              & 18.42                           \\
        FT+L1      & 21.07$\pm$10.79              & 7.86                           & 50.50$\pm$4.85              & 9.82                            \\
        GAGD       & 8.46$\pm$4.36                & 4.76                           & 54.82$\pm$4.67              & \textbf{5.50}                   \\
        GA         & 5.84$\pm$4.44                & 7.37                           & 48.10$\pm$6.01              & 12.23                           \\
        RL         & 44.60$\pm$4.18               & 31.39                          & 70.06$\pm$8.03              & 9.74                            \\
        SU      & 31.69$\pm$5.45               & 18.47                          & 91.55$\pm$1.17              & 31.23                           \\
        SU+ours & 10.53$\pm$2.21               & \underline{ 2.69}                     & 54.64$\pm$3.79              & \underline{ 5.68}                      \\ \hline
                   & \multicolumn{2}{c}{Setup3} & \multicolumn{2}{c}{Setup4} \\
                   & \multicolumn{1}{c}{MIA}      & \multicolumn{1}{c}{Gap}    & \multicolumn{1}{c}{MIA}     & \multicolumn{1}{c}{Gap}     \\
        Retrain    & 85.96$\pm$2.06               & \multicolumn{1}{l}{}           & 74.73$\pm$8.77              & \multicolumn{1}{l}{}            \\
        FT         & 25.97$\pm$17.71              & 59.99                          & 33.33$\pm$21.22             & 41.41                           \\
        FT+L1      & 43.45$\pm$13.83              & 42.51                          & 31.16$\pm$19.20             & 43.58                           \\
        GAGD       & 80.03$\pm$7.76               & \underline{ 5.93}                     & 29.59$\pm$17.55             & 45.14                           \\
        GA         & 67.11$\pm$6.66               & 18.85                          & 16.39$\pm$17.04             & 58.35                           \\
        RL         & 35.60$\pm$24.05              & 50.36                          & 53.73$\pm$25.17             & 21.00                           \\
        SU      & 98.02$\pm$0.82               & 12.06                          & 90.48$\pm$5.18              & \underline{ 15.74}                     \\
        SU+ours & 83.47$\pm$3.95               & \textbf{2.49}                  & 83.96$\pm$3.52              & \textbf{9.23}                 \\ \hline         
        \end{tabular}
        \caption{MIA and MIA gap in output probability}
        \label{tab:mia_prob}
        \end{table*}

        \begin{table*}[htbp]
        \centering
        \begin{tabular}{l|cc|cc} \hline
                   & \multicolumn{2}{c}{Setup1}      & \multicolumn{2}{c}{Setup2}     \\
                   & \multicolumn{1}{c}{MIA}      & \multicolumn{1}{c}{Gap}    & \multicolumn{1}{c}{MIA}     & \multicolumn{1}{c}{Gap}     \\
        Retrain & 18.91$\pm$0.65               & \multicolumn{1}{l}{}           & 48.36$\pm$1.13              & \multicolumn{1}{l}{}            \\
        FT      & 15.81$\pm$1.49               & 3.09                           & 41.74$\pm$1.14              & \textbf{6.62}                   \\
        FT+L1   & 17.77$\pm$1.30               & \textbf{1.14}                  & 3.90$\pm$0.49               & 44.46                           \\
        GAGD    & 12.25$\pm$1.29               & 6.66                           & 32.00$\pm$3.61              & 16.36                           \\
        GA      & 9.31$\pm$1.14                & 9.60                           & 4.47$\pm$0.51               & 43.89                           \\
        RL      & 61.40$\pm$3.35               & 42.50                          & 82.31$\pm$1.36              & 33.95                           \\
        SU      & 43.93$\pm$6.37               & 25.02                          & 93.93$\pm$0.30              & 45.57                           \\
        SU+ours & 15.89$\pm$1.22               & \underline{ 3.01}                     & 32.01$\pm$0.61              & \underline{ 16.34}                     \\
                          & \multicolumn{2}{c}{Setup3} & \multicolumn{2}{c}{Setup4} \\
                   & \multicolumn{1}{c}{MIA}      & \multicolumn{1}{c}{Gap}    & \multicolumn{1}{c}{MIA}     & \multicolumn{1}{c}{Gap}     \\
        Retrain & 62.26$\pm$0.81               & \multicolumn{1}{l}{}           & 59.29$\pm$1.05              & \multicolumn{1}{l}{}            \\
        FT      & 41.61$\pm$1.52               & 20.64                          & 42.71$\pm$3.12              & 16.58                           \\
        FT+L1   & 46.12$\pm$1.78               & \underline{ 16.14}                    & 45.63$\pm$2.52              & \underline{ 13.65}                     \\
        GAGD    & 41.90$\pm$1.47               & 20.36                          & 38.69$\pm$8.79              & 20.60                           \\
        GA      & 8.51$\pm$0.53                & 53.75                          & 8.16$\pm$3.43               & 51.12                           \\
        RL      & 81.63$\pm$1.40               & 19.37                          & 84.20$\pm$1.19              & 24.91                           \\
        SU      & 95.80$\pm$0.34               & 33.54                          & 88.26$\pm$0.91              & 28.97                           \\
        SU+ours & 47.07$\pm$1.02               & \textbf{15.19}                 & 49.04$\pm$0.21              & \textbf{10.24}                  \\ \hline         
        \end{tabular}
        \caption{MIA and MIA gap in entropy}
        \label{tab:mia_entr}
        \end{table*}

        \begin{table}[h]
            \centering
            {\fontsize{9}{11}\selectfont
            \begin{tabular}{c|ccc}
            \hline
             Setup & ours  & SU  & GAGD  \\ \hline
            1 & 2.99$\pm$0.01 & 9.98$\pm$0.02 & 18.05$\pm$0.02 \\
            2 & 3.44$\pm$0.08 & 29.94$\pm$0.09 & 54.87$\pm$0.12 \\
            3 & 4.45$\pm$0.42 & 71.26$\pm$0.26 & 132.47$\pm$0.38 \\
            4 & 4.36$\pm$0.04 & 37.74$\pm$0.02 & 69.47$\pm$0.13 \\ \hline
            \end{tabular}
            }
            \caption{The execution time of the strategy proposed by PeriUn and one epoch of SU and GAGD. The execution time of PeriUn is computationally negligible compared to the time required for a single epoch of unlearning methods such as SU and GAGD.}
            \label{tab:time}
            \end{table}

\begin{table}[h]
\centering
 {\fontsize{9}{11}\selectfont
\begin{tabular}{cccc}
\hline
Setup & $\mathcal{M}_r$ vs. $\mathcal{M}_{\text{base}}$ & $\mathcal{M}_r$ vs. $\mathcal{M}_{\text{ours}}$ & $\Delta$ \\ \hline
1     & 5.17$\pm$0.79                                   & 5.03$\pm$0.56                                   & 0.14     \\
2     & 27.46$\pm$1.56                                  & 22.26$\pm$0.56                                  & 5.19     \\
3     & 30.30$\pm$0.90                                  & 21.71$\pm$1.19                                  & 8.59     \\
4     & 34.21$\pm$1.86                                  & 24.69$\pm$0.89                                  & 9.53     \\ \hline
\end{tabular}
}
\caption{Percentage of disagreement in correct/incorrect predictions between $(\mathcal{M}_r, \mathcal{M}_{\text{base}})$ and $(\mathcal{M}_r, \mathcal{M}_{\text{ours}})$. 
A value of 100\% indicates complete disagreement (i.e., all predictions differ), while lower values indicate higher consistency between models. 
$\Delta$ denotes the mean difference between the two settings, showing that our method aligns more closely with the excluded samples than the baseline.}
    \label{tab:addi_quan_diff}
\end{table}

\end{document}